\documentclass[conference]{IEEEtran}
\newcommand{\myparagraph}[1]{\vspace{0.3\baselineskip}\noindent{\textbf{#1.}}~}

\usepackage[linesnumbered,ruled,vlined,boxed,noend]{algorithm2e}
\usepackage{booktabs}
\usepackage{amsmath}
\usepackage{amsthm}
\usepackage{amssymb}
\usepackage{graphicx}
\usepackage{enumitem}
\usepackage{multirow}
\usepackage{colortbl}
\usepackage{diagbox}
\usepackage{dsfont} 
\usepackage{mathrsfs}
\usepackage{algorithmic}
\usepackage{amssymb}
\usepackage{xcolor}
\usepackage{lipsum}
 
\newcommand\blfootnote[1]{%
  \begingroup
  \renewcommand\thefootnote{}\footnote{#1}%
  \addtocounter{footnote}{-1}%
  \endgroup
}

\setlist[itemize]{leftmargin=*}

\usepackage[colorlinks=true,linkcolor=red,urlcolor=black,citecolor=blue]{hyperref}

\DeclareMathOperator*{\argmin}{argmin}

\newcommand{\kmeans}{{{\texttt{$k$-means}}}\xspace}

\newcommand{\pick}{{{\texttt{Dask-means}}}\xspace}

\newcommand{\Tdrive}{{\small \textsf{T-drive}}\xspace}
\newcommand{\Porto}{{\small \textsf{Porto}}\xspace}
\newcommand{\ArgoAVL}{{\small \textsf{Argo-AVL}}\xspace}
\newcommand{\DRD}{{\small \textsf{3D-RD}}\xspace}
\newcommand{\ArgoPC}{{\small \textsf{Argo-PC}}\xspace}
\newcommand{\Shapenet}{{\small \textsf{Shapenet}}\xspace}
\newcommand{\ArgoETD}{{\small \textsf{Argo-ETD}}\xspace}
\newcommand{\ApollTD}{{\small \textsf{Apoll-TD}}\xspace}
\newcommand{\finding}{\noindent{\underline{\textit{Observations}}}.~}

\newcommand{\XGBoost}{{\small \texttt{XGBoost}}\xspace}
\newcommand{\DisNet}{{\small \texttt{DisNet}}\xspace}
\newcommand{\AutoML}{{\small \texttt{AutoML}}\xspace}

\newcommand{\ra}[1]{\renewcommand{\arraystretch}{#1}}

\newcounter{cN}
\definecolor{Maroon}{RGB}{128, 0, 0}

\usepackage{etoolbox}
\usepackage{marginnote}
\makeatletter
\let\oldmarginnote\marginnote
\renewcommand*{\marginnote}[1]{%
	\begingroup%
	\ifodd\value{page}
	\if@firstcolumn\normalmarginpar\fi
	\else
	\if@firstcolumn\else\normalmarginpar\fi
	\fi
	\oldmarginnote{\textcolor{brown}{#1}}%
		\oldmarginnote{}
	\endgroup%
}

\usepackage{color,soul}
\setul{0.5ex}{0.3ex}
\definecolor{Green}{rgb}{0,1,0}%{0,1,0}
\setulcolor{Green}

\definecolor{NavyBlue}{rgb}{0.0, 0.0, 0.5}

\ifCLASSOPTIONcompsoc
  \usepackage[nocompress]{cite}
\else
  \usepackage{cite}
\fi
\ifCLASSINFOpdf
\else
\fi  
\begin{document}
\title{On Simplifying Large-Scale Spatial Vectors: Fast, Memory-Efficient, and Cost-Predictable $k$-means}
%On Simplifying Large-Scale Spatial Vectors: Fast, Memory-Efficient, and Cost Predictable $k$-means
\author{\IEEEauthorblockN{Yushuai Ji\IEEEauthorrefmark{2}, Zepeng Liu\IEEEauthorrefmark{2}, Sheng Wang\IEEEauthorrefmark{2}\IEEEauthorrefmark{1}, Yuan Sun\IEEEauthorrefmark{4}, and Zhiyong Peng\IEEEauthorrefmark{2}\IEEEauthorrefmark{3}}
\IEEEauthorblockA{\IEEEauthorrefmark{2}School of Computer Science, Wuhan University \\ 
\IEEEauthorrefmark{3}Big Data Institute, Wuhan University \\ 
\IEEEauthorrefmark{4}La Trobe Business School, La Trobe University}
\IEEEauthorblockA{[yushuai, liuzp\_063, swangcs, peng]@whu.edu.cn, yuan.sun@latrobe.edu.au}
}

%
%A Unified Exact Searching Framework over Multi-source Vectors in Autonomous Vehicles
%IEEE TRANSACTIONS ON KNOWLEDGE AND DATA ENGINEERING
\markboth{IEEE TRANSACTIONS ON KNOWLEDGE AND DATA ENGINEERING,~Vol.~X, No.~X, MAY~2024}%
{Shell \MakeLowercase{\textit{et al.}}: Bare Demo of IEEEtran.cls for Computer Society Journals}
\maketitle
\begin{abstract}
The \kmeans algorithm can simplify large-scale spatial vectors, such as 2D geo-locations and 3D point clouds, to support fast analytics and learning. However, when processing large-scale datasets, existing \kmeans algorithms have been developed to achieve high performance with significant computational resources, such as memory and CPU usage time. These algorithms, though effective, are not well-suited for resource-constrained devices.
In this paper, we propose a fast, memory-efficient, and cost-predictable \kmeans called \pick. We first accelerate \kmeans by designing a memory-efficient accelerator, which utilizes an optimized nearest neighbor search over a memory-tunable index to assign spatial vectors to clusters in batches.
We then design a lightweight cost estimator to predict the memory cost and runtime of the \( k \)-means task, allowing it to request appropriate memory from devices or adjust the accelerator's required space to meet memory constraints, and ensure sufficient CPU time for running \( k \)-means. Experiments show that when simplifying datasets with scale such as $10^6$, \pick uses less than $30$MB of memory, achieves over $168$ times speedup compared to the widely-used Lloyd's algorithm. We also validate \pick on mobile devices, where it demonstrates significant speedup and low memory cost compared to other state-of-the-art (SOTA) \kmeans algorithms. Our cost estimator estimates the memory cost with a difference of less than $3\%$ from the actual ones and predicts runtime with an MSE up to $33.3\%$ lower than SOTA methods.
%The experimental results show that when used to simplify point clouds' scale from $10^7$ to $10^4$, our \pick algorithm only uses less than 500MB memory and can achieve up to $1000\times$ speedup over the widely-used Lloyd's algorithm. 

% \kmeans clustering has been used to simplify and compress large-scale point clouds by reducing $n$ points to $k$ centroids.
% %
% Most \kmeans accelerating algorithms need to maintain up to $k$ centroid bounds for each point to prune unnecessary distance computations. 
% %
% For large $n$ and $k$, these algorithms cannot finish the clustering due to slow convergence or by running out of memory.
% %
% We propose an algorithm called \pick for fast and memory-efficient point cloud \kmeans clustering. 
% %\jf{need to better articulate the following sentence}
% \sheng{We build lightweight indexes on both points and centroids, and leverage an optimized nearest centroid search algorithm to efficiently assign points to centroids in batch.
% The index size can be auto-configured according to the available memory on devices.}
% Experiments show that we can easily achieve $1000\times$ speedup over the widely known Lloyd's algorithm when $n\ge10^7$ and $k\ge10^4$.
\end{abstract}

\IEEEdisplaynontitleabstractindextext
\IEEEpeerreviewmaketitle
\section{Introduction}
\label{sec:intro}
%移动端产生大量的数据集
%数据集要cluster
Sensors, such as GPS and lidar, are commonly found in resource-constrained devices like autonomous vehicles (AVs) \cite{WangBCC21} and drones \cite{Shao2020,Guo2020a}. They generate a wealth of spatial vectors \cite{He0KXJ0022}, which are geometric representations of spatial objects, for example, 2D trajectory datasets and 3D point cloud datasets collected from GPS and lidar deployed in AVs \cite{Chang2019}. \blfootnote{$^{\ast}$Sheng Wang is the corresponding author.}

These spatial vectors can be widely applied on resource-constrained devices in visualization and learning tasks such as classification \cite{Lang2019} and segmentation \cite{Qi2017, Hu2020}.
For technologies such as 3D object recognition \cite{LvLZ21}, processing all cloud points is unnecessary since a dense point cloud contains many redundant spatial vectors, and processing all spatial vectors significantly increases storage and processing costs.
%The most straightforward way to tackle this limitation is simplifying a dataset by randomly selecting a subset of spatial vectors from it \cite{Hu2020}. However, as shown in Fig.~\ref{fig:motivation}, randomly selected spatial vectors may not be evenly distributed and, therefore, may not correctly represent the dataset properly \cite{Roynard2018}. Clustering algorithms (e.g., \kmeans \cite{Lloyd1982}) are more effective alternatives (also shown in Fig.~\ref{fig:motivation}), and they have been widely used to simplify point clouds \cite{Sun2019c, Mariam2020, Xu2020, Yin2020} and {summarize datasets} \cite{Kleindessner2019, Mohiuddin2019, Chirigati2021}.

\begin{figure}
	\centering
	\includegraphics[width=9cm]{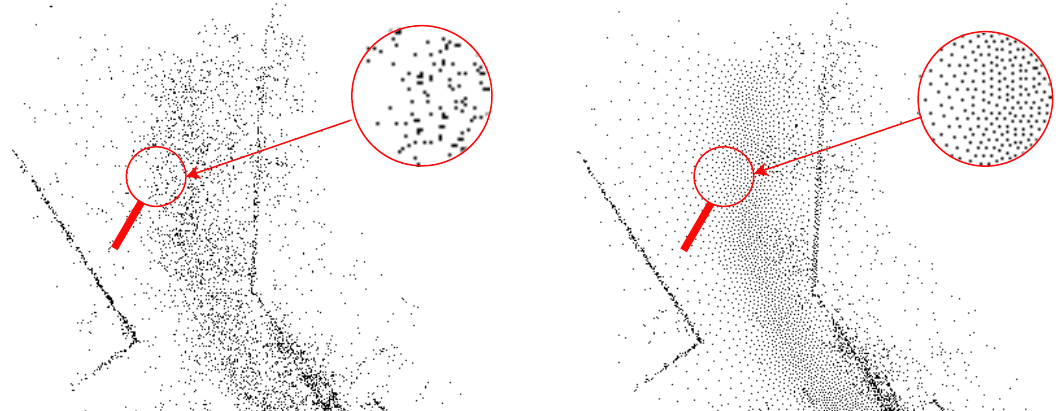}
	\vspace{-1.5em}
	\caption{The simplified point clouds by random sampling (left) and {our \kmeans clustering algorithm} (right).}
	\vspace{-2em}
	\label{fig:motivation}
\end{figure}

The most straightforward way to tackle this limitation is by simplifying a dataset, and the two most widely used simplifying methods are random selection \cite{Hu2020} and clustering algorithms (e.g., \kmeans \cite{Lloyd1982}). However, randomly selecting a subset of spatial vectors from the dataset may not be evenly distributed and, therefore, cannot accurately represent the dataset \cite{Roynard2018}. This makes \kmeans a better choice and widely used to simplify point clouds \cite{Sun2019c, Mariam2020, Xu2020, Yin2020} and summarize datasets \cite{Kleindessner2019, Mohiuddin2019, Chirigati2021}. For example, as shown in Fig.~\ref{fig:motivation}, the spatial vectors selected by our \kmeans algorithm are more evenly distributed than the randomly selected ones.

However, as the scales of datasets expand significantly and reach millions, technologies such as object detection in AVs \cite{LvLZ21} still require the \kmeans algorithm to be highly efficient. Existing \kmeans algorithms have been developed to achieve efficiency using significant computational resources, such as memory. These algorithms, though effective, are not well-suited for resource-constrained devices. For example, Google Coral \cite{FlyingFox} and Raspberry Pi \cite{Brand2018} typically have 4GB of memory. This raises a critical research question: \textit{how to design a fast and memory-efficient \kmeans algorithm to simplify large-scale spatial vectors on resource-constrained devices?}

To answer the question, we need to tackle two key challenges: 1) high space cost for storing bounds and indexes to reduce unnecessary distance computations for accelerating $k$-means tasks, and 2) degradation of the efficiency of \kmeans algorithms due to limited memory resources for storing information, such as indexes, and insufficient CPU resources for running it. 
% limited resources impair the efficiency of \kmeans algorithms by constraining memory usage, which is inadequate for storing information, such as indexes, for acceleration, and by providing insufficient CPU usage time for running it. 
Various techniques have been proposed to address these challenges, but they still have shortcomings when applied to resource-constrained devices, as summarized below.

%框架图说明
\myparagraph{Sacrificing Substantial Space for Accelerating} Existing \kmeans algorithms have high time or space complexity, making them inapplicable to simplify large-scale datasets on resource-constrained devices. Memory-efficient \kmeans algorithms such as {Lloyd's algorithm} \cite{Lloyd1982}, \texttt{Index} \cite{Moore2000}, \texttt{Hamerly} \cite{Hamerly2010}, and \texttt{NoBound} \cite{Xia2020}, are computationally slow, especially for clustering tasks that involve both a large number of spatial vectors and clusters. Although there are algorithms that trade off memory for speed (e.g., \cite{Elkan2003, Newling2016, Rysavy2016a}), they require more memory than is available on these devices.

\myparagraph{Inaccruate Memory \& Runtime Estimation} 
%It is challenging to accurately predict the resources, including memory cost and CPU usage time, consumed by \kmeans. 
For memory cost estimation, existing methods are either designed for programs like Java-like projects \cite{AlbertGG10}, which cannot be directly applied to \kmeans, or are tailored for a few machine learning (ML) models \cite{TensorFlow, CanzianiPC16, GaoLZLZLY20}, such as neural networks. For runtime estimation, most methods \cite{DoanK17, EggenspergerLH18, TueroB21} involve training an ML model to predict runtime based on features provided by \kmeans. However, existing methods often lead to high training overhead because the models require a large number of samples to be generated for model training, which also requires substantial computational time.

%data simplification $k$-Means algorithm fmc
In this paper, we propose \pick, a fast, memory-efficient, and cost-predictable \underline{da}taset \underline{s}implification \underline{$k$-means} algorithm for large-scale spatial vectors. To accelerate Lloyd's algorithm without costing substantial memory, we build indexes on spatial vectors and cluster centroids. The index supports optimized $\mathtt{k}$ Nearest Neighbor ($\mathtt{k}$NN\footnote{We use $\mathtt{k}$ to differentiate from $k$ in $k$-means as they represent different concepts.}) search to assign spatial vectors to a cluster efficiently. To predict the memory cost and runtime of \pick accurately, we propose a lightweight cost estimator, which can analyze the space cost of the pruning mechanism, and estimate the overall runtime by predicting the iteration number and runtime of each iteration separately. Overall, our main contributions are: 
\begin{itemize}
     \item We design pruning mechanisms that apply a three-pronged optimized $\mathtt{k}$NN search on the centroid index to batch prune spatial vectors to accelerate $k$-means tasks (see Section~\ref{sec:pick}).

    \item We predict the memory cost of $k$-means tasks by building a mapping function between the dataset and the index, and estimate runtime by separately predicting the iteration number and the runtime of each iteration (see Section~\ref{sec:costmodel}).
	
   \item Experiments on the tested datasets show that \pick accelerates Lloyd's algorithm by up to 168 times. Our cost estimator predicts memory cost with a difference of less than 3\% from the actual values and estimates runtime with an MSE 33.3\% lower than SOTA models (see Section~\ref{sec:exp}).
\end{itemize}
%1

\section{Background and Preliminaries}
\label{sec:pre}

\subsection{Notations}
We use different text formatting styles to represent mathematical concepts: plain letters for scalars, bold letters for vectors, capitalized letters for objects, and bold capitalized letters for a set containing vectors. 
For example, $x$ stands for a scalar, $\mathbf{p}$ represents a spatial vector, $N$ denotes an index node, and $\mathbf{D}$ represents a dataset. Without loss of generality, we denote the $d$-dimensional Euclidean space as $\mathbb{R}^{d}$, the set of positive real numbers as $\mathbb{R}^+$, and the set of positive integers as $\mathbb{Z}^{+}$. Moreover, we use $||\cdot||$ as the Euclidean norm. The notation details are presented in Table~\ref{tab:notations}.
\begin{table}[t]
%\ra{1.3}
\centering
\setlength{\tabcolsep}{6pt}
%\setlength{\abovecaptionskip}{1em} 
%\vspace{-1em}
\caption{Summary of notations.}
\label{tab:notations}
\vspace{-1em}
\scalebox{1.03}{\begin{tabular}{cc}  
\toprule   
\textbf{Notation} & \textbf{Description}  \\  
\midrule
$n \in \mathbb{Z}^{+}$ &  The dataset size \\
$\mathbf{p} =(x_1,x_2,...,x_d) \in \mathbb{R}^{d}$ &  The spatial vector\\
$\mathbf{D}=\{\mathbf{p}_i\}_{i=1}^{n} \in \mathbb{R}^{n \times d}$ &  The dataset\\
$k \in \mathbb{Z}^{+}$ &  The number of clusters \\
$S =\{S_1, S_2, \cdots, S_k\}$ &  $k$ exclusive subsets \\
$\mathbf{c}_j \in \mathbb{R}^d$ &   The mean of the spatial vectors in $S_j$ \\
$N$ &  The spatial vector index node (ball node)\\
$\mathbf{p}^*$ & $\mathbf{p}^*$ is the pivot of a node $N$ \\
$r$ &  The radius of $N$ \\
$N_C$ &  The centroids index node \\
$m \in \mathbb{R}$ &  The available memory \\
$f \in \mathbb{Z}^{+}$ & The leaf node capacity\\
$t \in \mathbb{R}^{+}$ &  The runtime of \kmeans\\
$q \in \mathbb{Z}^{+}$ &  The maximum number of iterations\\
\bottomrule  
\end{tabular}}
\vspace{-1.8em}
\end{table}

\subsection{Definition of $k$-means}
The $k$-means is a bivariate optimization problem. Given a dataset $\mathbf{D}=\{\mathbf{p}_1, \mathbf{p}_2, \cdots, \mathbf{p}_n\} \in \mathbb{R}^{n \times d}$, $k$-means aims to partition $\mathbf{D}$ into $k$ exclusive subsets $S=\{S_1, S_2, \cdots, S_k\}$ to minimize the \textit{Sum of Squared Error}:
\begin{equation}
\argmin\limits_{S} \sum_{j=1}^{k}\sum_{\mathbf{p}\in S_j}\|\mathbf{p}-\mathbf{c}_j\|^2,
\end{equation}

\noindent where the \emph{centroid} $\small \mathbf{c}_j=\frac{1}{|S_j|}\sum_{\mathbf{p}\in S_j}\mathbf{p}$ is the mean of spatial vectors in cluster $S_j$. Lloyd's algorithm \cite{Lloyd1982} is one of the most widely used methods to solve the $k$-means problem by assigning each spatial vector to its nearest centroid and iteratively refining the centroids. The algorithm requires computing $n \times k$ distances in the assignment phase of each iteration, which is computationally prohibitive for applying it to datasets where both $n$ and $k$ are large. 
%On the other hand, Lloyd's algorithm does not require any additional memory (see Section~\ref{sec:exch}), so it is the most memory-efficient and the only clustering algorithm available in the Point Cloud Library~\cite{Rusu2011,pcl}.

% The Lloyd's algorithm \cite{Lloyd1982} computes \kmeans by assigning every point to their nearest centroid through scanning $k$ times; hence it needs $n \times k$ computations in the assignment of each iteration, which is not feasible for clustering point cloud data with $k\ge10^4$.
%
%\jf{what is a large k?}
%
% However, it only needs to load the dataset and centroids and does not have other costs, so it is the most memory-efficient algorithm, and
% the Point Cloud Library (PCL) \cite{Rusu2011} implemented Lloyd's algorithm as the only \kmeans algorithm.\footnote{\url{https://github.com/PointCloudLibrary/pcl}}

%\jf{Is Moore's index a variation of ball tree? If so, you don't need to mention kd-trees -- it is confusing here}
%
%\jf{you should not say "The Lloyd's algorithm"}
%
\subsection{Accelerated Lloyd's Algorithms for \kmeans}
\label{sec:exch}
We focus on techniques such as hardware-based algorithms, index-based algorithms, and sequential algorithms to speed up $k$-means, as they yield the same results as Lloyd's algorithm.
%Various techniques have been developed to accelerate {Lloyd's algorithm}, e.g., building an index for spatial vectors so that they can be assigned in batch, or designing upper/lower bounds on the distance between spatial vectors and clusters' centroids to avoid unnecessary distance computations. These techniques can generate the same results faster but at the expense of additional space.

%There are other approximate fast Lloyd's algorithms, such as mini-batch \cite{Sculley2010,Newling2016a} which samples a subset of spatial vectors first and then applies Lloyd's algorithm on the spatial vector samples. In this sense, the mini-batch may generate different clustering results compared to Lloyd's algorithm, even given the same initialization.
\myparagraph{Hardware-based Algorithm}Several researchers \cite{KrulisK20,Li2013} design parallel \kmeans algorithm for GPUs. For instance, Li et al. \cite{Li2013} develop a parallel \kmeans using a general-purpose parallel programming model. Although the methods are applicable to edge devices, they require significant computational resources, making them unsuitable for resource-constrained devices. Others focus on accelerating \kmeans on specific processors, such as CPU-FPGA \cite{Abdelrahman20}, FPGA \cite{WangGJZ21}, and heterogeneous many-core supercomputers \cite{DYuZLJYWFYT20}. Additionally, Bender et al. \cite{Bender2015} use two-level memory systems to speed up $k$-means. However, they lack generality, as many edge devices do not have this type of processor or storage system.

\begin{figure}
	\centering
	\includegraphics[width=8.7cm]{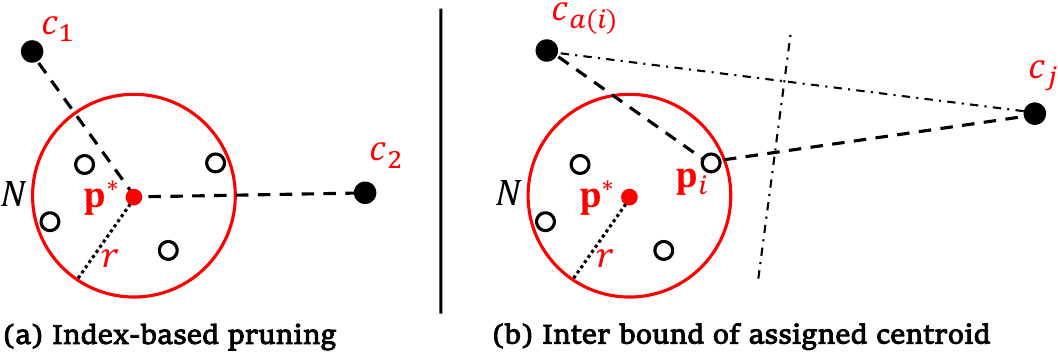}
	\vspace{-0.5em}
	\caption{Pruning using ball node and inter bound.}
	\vspace{-1.5em}
	\label{fig:balls}
\end{figure}

\myparagraph{Index-based Algorithm} Instead of assigning spatial vectors one by one, Moore et al. \cite{Moore2000} proposed an index that stores spatial vectors in a hierarchical tree structure called \textit{Ball-tree} \cite{Omohundro1989}. The \textit{spatial vector index} can avoid the distance computation between a centroid and a set of spatial vectors. For example, given two centroids $\mathbf{c}_1$ and $\mathbf{c}_2$,
all the spatial vectors in a ball node $N$ are closer to centroid $\mathbf{c}_1$ than $\mathbf{c}_2$ if
\begin{equation}
\label{equ:ball}
\|\mathbf{p^*}-\mathbf{c}_1\| + r < \|\mathbf{p^*}-\mathbf{c}_2\| - r,
\end{equation}
where $\mathbf{p}^*$ is the pivot of a node $N$ that bounds all spatial vectors within a radius $r$, as shown in Fig.~\ref{fig:balls}(a). The index structure requires extra memory cost, which is proportional to the number of nodes in the Ball-tree. The drawback of the index-based algorithm is that it scans all cetroids one by one, which needs $k$ distance computations, to assign the spatial vectors in an index node to their nearest centroid in batch.

%$\mathbf{p}_i$ is a spatial vector, $a(i)$ denotes the id of the centroid that $\mathbf{p}_i$ was assigned in the previous iteration, $c_1$, $c_2$, $c_j$, and  $c_{a(i)} $are centroids.

%The drawback of the index-based algorithm is that it needs $k$ distance computations to assign the spatial vectors in an index node to their nearest centroid in batch. In Section~\ref{sec:pick}, \pick overcomes this limitation by introducing a memory-tunable index for both the centroids and spatial vectors, and uses $\mathtt{k}$NN to accelerate the assignment process.

%and use a \textit{nearest neighbor search algorithm} to accelerate this process (see Section~\ref{sec:pruning}).

%\sheng{The main drawback of this index-based algorithm is that it needs $k$ times of distance computations to find the nearest centroid and assign for every point and node.} \songsong{To overcome this drawback, we build lightweight indexes on $k$ centroids and use a \textit{nearest neighbor search algorithm} to accelerate this process (see Section~\ref{sec:pruning}).}
%%***没看***Sequential Algorithms
\myparagraph{Sequential Algorithms}As shown in Fig.~\ref{fig:balls}(b), to check whether a spatial vector $\mathbf{p}_i$ belongs to a cluster with centroid $\mathbf{c}_j$, Elkan et al. \cite{Elkan2003} store the lower bound on the distance from $\mathbf{p}_i$ to $\mathbf{c}_j$, which needs $O(nk)$ memory for all pairs of spatial vectors and centroids. Firstly, an \textit{inter bound} is derived as $\|\mathbf{c}_{a(i)}-\mathbf{c}_j\|$, where $a(i)$ denotes the id of the centroid that $\mathbf{p}_i$ was assigned in the previous iteration.\footnote{To facilitate our illustration, each cluster is simply represented by its centroid if no ambiguity is caused.} If $\|\mathbf{p}_i-\mathbf{c}_{{a(i)}}\|< \|\mathbf{c}_{a(i)}-\mathbf{c}_j\| / 2$, then $\mathbf{c}_j$ can be pruned.
As storing all bounds to every centroid uses much space, Hamerly et al. \cite{Hamerly2010} proposed to choose the minimum one as the only centroid inter bound:
\begin{equation}
\mathbf{cb}[a(i)] = \min_{\mathbf{c}_j\in \mathbf{C} \& j\ne a(i)} \|\mathbf{c}_{a(i)}-\mathbf{c}_j\|.
\label{equ:inter}
\end{equation}

%Another bound proposed by 
Elkan et al. \cite{Elkan2003} stored the computed distances to accelerate the next iteration, and computes the moving distance of each centroid $\Delta[j] = \|\mathbf{c}_{j}-\mathbf{c}_j^{'}\|$, also called \textit{drift}, to estimate the lower bound using triangle inequality: $\|\mathbf{p}_i - \mathbf{c}_{j}\| \ge \|\mathbf{p}_i-\mathbf{c}_j^{'}\| - \|\mathbf{c}_{j}-\mathbf{c}_j^{'}\|$, where $\mathbf{c}_j^{'}$ denotes the position of centroid $j$ in the previous iteration.
Drake \cite{Drake2013}, Hamerly and Drake \cite{Hamerly2015a}, Newling and Fleuret \cite{Newling2016b}, Ryšavý
and Hamerly \cite{Rysavy2016a} proposed even tighter bounds, but they all require substantial memory and become prohibitively costly when $k$ is large.
Furthermore, since the bounds must be updated across iterations, this overhead slows down the clustering process, making these algorithms unsuitable for simplifying large-scale point clouds. %, and we will not further elaborate on them. 

% in this paper.

%to cluster point clouds, so we will not elaborate on them in this paper.

%\subsection{Memory-Efficient Algorithms}
Several memory-efficient sequential algorithms were proposed, including \cite{Hamerly2010} described above; \cite{Drake2013} that stores $\frac{k}{4}$ minimum lower bounds; Yinyang \cite{Ding2015} that divides $k$ centroids into $\frac{k}{10}$ groups and each group has one lower bound; and Dual-tree \cite{Curtin2017} that extends the single upper and lower bound of \cite{Hamerly2010} to index-based algorithm \cite{Moore2000}, and use index to group centroids \cite{Ding2015} for pruning centroids in batch. In contrast to \cite{Hamerly2010}, Dual-tree \cite{Curtin2017} needs extra memory for a spatial vector index, where each spatial vector and node maintains two bounds and other pruning information. %such as the number of pruned centroids.

% \cite{Drake2013} by sorting $b=k/4$ minimum lower bounds; Yinyang \kmeans \cite{Ding2015} divides $k$ centroids into $k/10$ groups, each group has one lower bound; Dual-tree \cite{Curtin2017} extends the single upper and lower bound of \cite{Hamerly2010} to index-based algorithm \cite{Moore2000}, and use index to group centroids \cite{Ding2015} for pruning centroids in batch.
% Different from \cite{Hamerly2010}, Dual-tree \cite{Curtin2017} needs extra memory on a heavy point index, where each point and node maintains two bounds and other pruning information such as the number of pruned centroids.

Moreover, Xia et al \cite{Xia2020} accelerate \texttt{$k$-means} with no bound (short as NoBound), but it needs to create a centroid distance matrix ($k \cdot k$) in each iteration for pruning centroids outside a radius range. For an overview of all existing accelerating algorithms, we suggest readers refer to Section 4.2 of our recent evaluation paper \cite{WangUnik2020}. As shown in Table~\ref{tab:memory}, we compare existing memory-efficient algorithms from five new perspectives. %in four aspects.
%\pick reduces memory cost by maintaining no bounds across iterations with a memory-tunable index and assigns spatial vectors in batch without scanning all $k$ centroids.

%We will compare the efficiency with them in the experiments.
%All of them cannot work well when k is big.
%All of them need to maintain the bounds, which will also increase the memory cost.
%In a \kmeans evaluation, \cite{WangUnik2020} pointed out that frequent updates on bounds will also lead to slow performance of \kmeans.
%This is also slow to compute in the first iteration.
%We will study the memory costs in the clustering.
%\vspace{-1em}
% \usepackage{multirow}
\begin{table}[]
\ra{1.2}
\centering
\setlength{\tabcolsep}{3pt}
%\setlength{\abovecaptionskip}{1em} 
%\vspace{-1em}
%表2利用了啥
\caption{Comparison with memory-efficient sequential algorithms.}
\label{tab:memory}
\vspace{-1em}
\scalebox{0.97}{
\begin{tabular}{cccccc}
\toprule
\multirow{3}{*}{\textbf{Algorithm}}          & \textbf{Prune}      & \textbf{Update-}       & \textbf{Assign}       & \textbf{Memory-}     & \textbf{Run-}       \\
                                             & \textbf{Centroids}       & \textbf{free}         & \textbf{Points}       & \textbf{cost}     & \textbf{time}     \\
                                             & \textbf{In-batch}         & \textbf{Bounds}       & \textbf{In-batch}     & \textbf{Tunable}      & \textbf{Predictable}   \\  \midrule
\texttt{Hamerly} \cite{Hamerly2010}     & \checkmark & \texttimes                    & \texttimes                        & \texttimes           & \texttimes                     \\ 
\texttt{Drake} \cite{Drake2013}      &\checkmark   & \texttimes                    & \texttimes                        & \checkmark           & \texttimes                     \\
\texttt{Yinyang} \cite{Ding2015}     & \checkmark  & \texttimes                    & \texttimes                        & \checkmark            & \texttimes                    \\
\texttt{Dual-tree} \cite{Curtin2017} &\checkmark  & \texttimes                    & \checkmark                        & \texttimes             & \texttimes                  \\
\texttt{NoBound} \cite{Xia2020}      & \texttimes & \checkmark                    & \texttimes                        & \texttimes             & \texttimes                   \\ \midrule
\pick                         & \checkmark & \checkmark                    & \checkmark                        & \checkmark           & \checkmark                \\ \bottomrule
\end{tabular}}
\vspace{-2em}
\end{table}

\subsection{Cost Estimator}
\label{sec:cost_pre}
\myparagraph{Memory Cost Estimation} Several technologies \cite{VerbauwhedeSR94, AlbertGG10, HeoOY19} have been proposed to predict the memory cost for programs such as Java-like programs \cite{AlbertGG10}, and can be applied in \kmeans. For example, Verbauwhede et al. \cite{VerbauwhedeSR94} estimate the memory cost of digital signal processing programs by modeling array dependencies and execution sequences as an integer linear programming (ILP) problem, which is then solved using an ILP solver. Albert et al. \cite{AlbertGG10} introduce a parametric technology to infer the memory cost of Java-like programs by analyzing object lifetimes. Heo et al. \cite{HeoOY19} propose a resource-aware, flow-sensitive analysis towards estimating memory cost using online abstraction coarsening. However, they can only predict the memory cost for \kmeans written in specific programming languages.

\textit{TensorFlow} \cite{TensorFlow} and several ML model performance analysis works \cite{CanzianiPC16} estimate memory cost by summarizing the parameters, dataset, and outputs. However, they are just a subset of the whole memory cost. Moreover, TensorFlow cannot analyze the memory costs related to indexes and bounds, which can affect the final memory cost. Additionally, Gao et al. \cite{GaoLZLZLY20} propose DNNMem, which calculates the memory cost of the computation graph and the deep learning (DL) model runtime. However, it only works for DL models.

\myparagraph{Runtime Estimation}%\cite{KapusC19, TueroB21, EggenspergerLH18}
A bunch of models have been proposed for estimating runtime, but they are time-consuming. Models like Bayes DistNet \cite{TueroB21}, XGBoost \cite{GunnarssonBW23}, and AutoML \cite{MohrWTH21} require an impractical number of training samples before they can positively impact prediction time. Generating tens of thousands of $k$-means samples for training could take several hours or even days for a resource-constrained device. Eggensperger et al. \cite{EggenspergerLH18} propose DisNet to predict the runtime accurately by a neural network. However, these ML models repeatedly perform forward propagation, loss calculation, and backpropagation over multiple epochs until the neural network reaches satisfactory performance, which is time-consuming.

Alternative methods \cite{HutterXHL15,MohrWTH21,Leyton-BrownNS09,Yuping17} for predicting $k$-means runtime use linear regression, which is time efficient. For example, Leyton-Brown et al. \cite{Leyton-BrownNS09} use ridge regression to predict runtime. Fan et al. \cite{Yuping17} predict runtime using linear regression by data censoring. 
Hutter \cite{HutterXHL15} and Mohr \cite{MohrWTH21} compare various regression models in terms of training time, prediction time, and prediction error. However, their analysis applies to general ML models and shows low accuracy when predicting for the $k$-means tasks (see Section~\ref{sec:exp}).

Moreover, several models \cite{Yuping17,TangDBL10} use the posterior information during task execution to adjust the predicted runtime. For example, Fan et al. \cite{Yuping17} propose TRIP to reduce underestimation rates of prediction by incorporating elastic net regularization (two penalties) into the linear regression model. Similarly, Tang et al. \cite{TangDBL10} improve runtime accuracy by multiplying a user-supplied runtime estimate with an adjusting parameter. However, adding penalties or an adjusting parameter requires extensive experimentation to find the optimal values, which is not practical when the dataset or setting parameters in ML models change. 

\section{Framework of Dask-means}
\label{sec:framework}
\begin{figure}
	\centering
  \includegraphics[width=.48\textwidth]{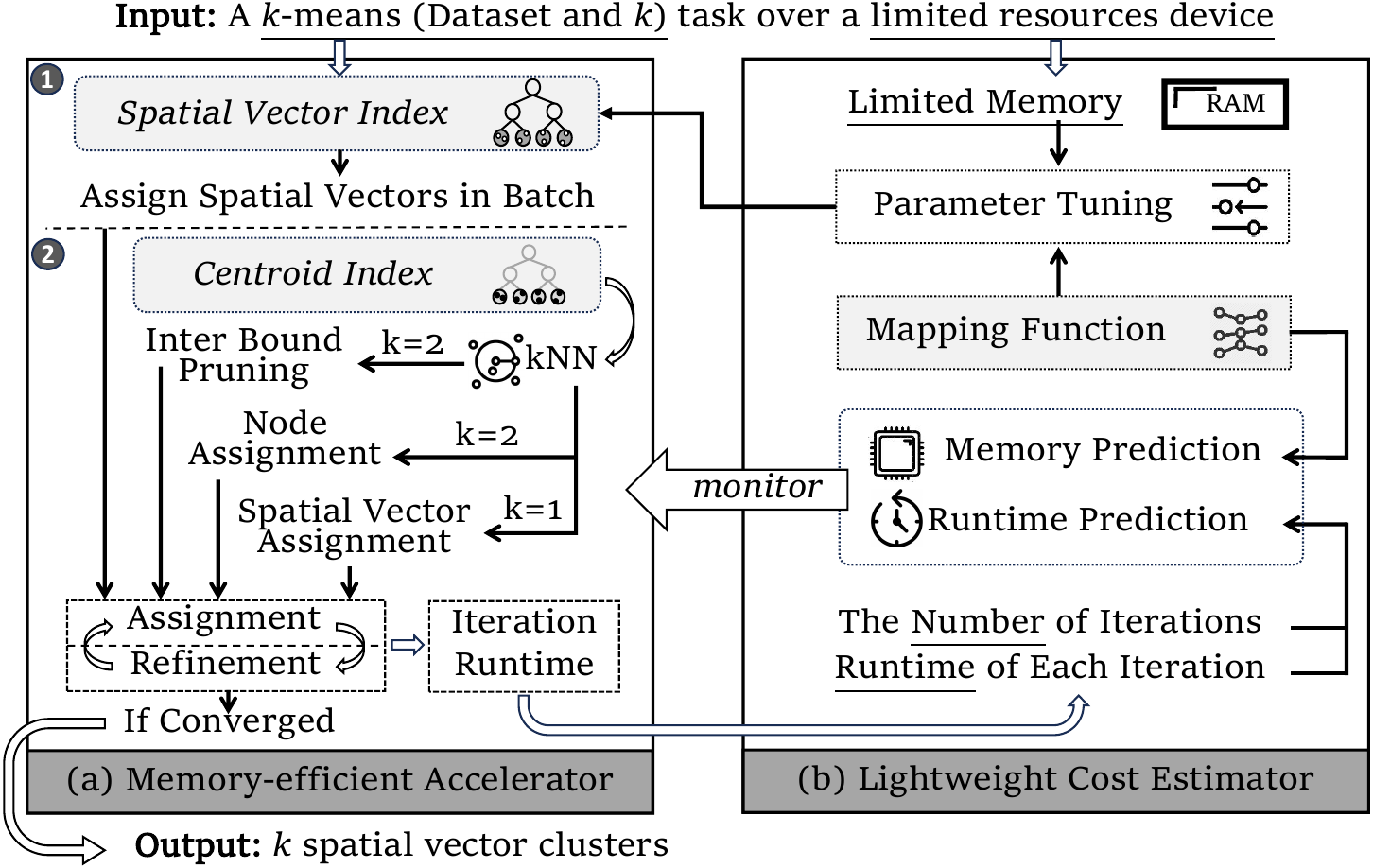}
	\vspace{-1em}
	\caption{Framework of \pick.}
	\label{fig:overview_dask}
	\vspace{-2em}
\end{figure}

%\pick accelerates the $k$-means on resource-constrained devices. We first design a memory-efficient accelerator to avoid unnecessary computations. We propose a cost model to accurately predict the runtime and memory for requesting devices to provide suitable resources, tuning the pruning mechanism parameter to accelerate $k$-means based on the given resource, such as memory.
%proposing a cost model to accurately predict time and memory cost w
In this section, we introduce \pick, which can accelerate $k$-means significantly, especially on resource-constrained devices. As shown in Fig.~\ref{fig:overview_dask}, \pick consists of two modules described below.

\myparagraph{Memory-efficient Accelerator}Recall that the index-based algorithm \cite{Moore2000} requires a time-intensive scan of all cluster centroids when assigning spatial vectors to clusters. Sequential algorithms \cite{Elkan2003, Drake2013}, in contrast, require substantial memory to maintain bounds for spatial vectors and nodes across iterations for centroid pruning. To avoid these limitations, as shown in Fig.~\ref{fig:overview_dask}(a), 
we construct an index on spatial vectors, denoted as spatial vector index. This index leverages nodes to represent a group of spatial vectors, thus avoiding distance computations between a centroid and a batch of spatial vectors. Then, we design an optimized $\mathtt{k}$NN search over the index built on centroids (namely the centroid index) and maintain the proposed inter bound to prune unnecessary computations.

%As shown in Fig.~\ref{fig:overview_dask}(a), we build indexes over spatial vectors, denoted as spatial vector index, and clusters' centroids, called centroid index. We design an optimized nearest centroid search method (referred to as $\mathtt{k}$NN) \footnote{We use $\mathtt{k}$ to differentiate from $k$ in $k$-means as they represent different concepts.} to accelerate node and spatial vector assignments.

\myparagraph{Lightweight Cost Estimator}
As shown in Fig.~\ref{fig:overview_dask}(b), we propose a lightweight cost estimator to predict the memory cost and the runtime for $k$-means. We first propose a memory estimate method to predict memory cost by building a mapping function between the dataset and the index. This method also allows us to adjust the hyperparameters of the proposed accelerator to build a memory-tunable index that accelerates the $k$-means task. We estimate the runtime by separately predicting the iteration number and each iteration's runtime. Notably, we then extract posterior information from the last iteration of the $k$-means task to adjust the predicted runtime.

\section{Memory-efficient Accelerator}
\label{sec:pick}
%制造割裂感
\subsection{Pruning Mechanisms}
\label{sec:pruning}
We design pruning mechanisms that apply a three-pronged optimized $\mathtt{k}$NN over the centroid index to batch prune nodes and spatial vectors, thus accelerating $k$-means without the need to store bounds for spatial vectors. We first prune distance computations between centroids and a set of spatial vectors by applying $\mathtt{k}$NN to find the nearest centroids of an index node (or a spatial vector), with the $\mathtt{k}$NN bounds inherited from parent nodes. We then use $\mathtt{k}$NN to search for the nearest centroids of the target one to avoid scanning all centroids. Notably, we add two drifts in estimating the inter bound to accelerate $\mathtt{k}$NN. 
%句子简单一点

%elaborate on the above pruning mechanisms progressively.}

\begin{figure}
	\centering
	\includegraphics[width=9cm]{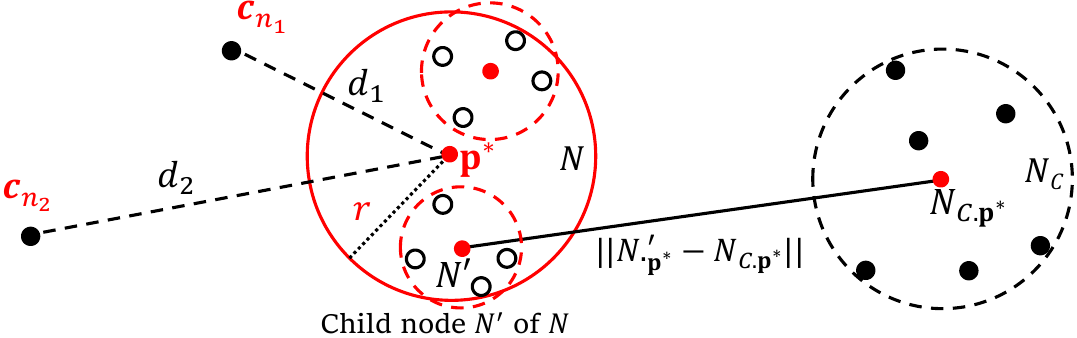}
	\vspace{-1.5em}
	\caption{Pruning with a single indexing tree, where a spatial vector node \( N \) contains two child nodes; pruning with centroid index node \( N_C \), where \( \mathbf{c}_{n_1} \) and \( \mathbf{c}_{n_2} \) represent the two nearest centroids to \( N \)'s pivot (\( \mathbf{p}^* \)), with the corresponding distances \( d_1 \) and \( d_2 \) (where \( d_2 > d_1 \)); \( N.\mathbf{p}^* \) refers to the pivot of \( N' \) and \( N_C.\mathbf{p} \) refers to the pivot of \( N_C \).}
	\vspace{-2em}
	\label{fig:pruning}
\end{figure}

\myparagraph{Indexes on Spatial Vectors and Centroids}
We build a Ball-tree index structure for spatial vectors with the root node denoted as $R$, and another Ball-tree index structure for the clusters' centroids with the root node denoted as $R_C$. {In Fig.~\ref{fig:pruning}, we show a toy example where an index node of spatial vectors (the big red circle on the left) covers its two child nodes (the small red dotted circles); the black circle on the right denotes a centroid node which covers six centroids.} Note that the Ball-tree for spatial vectors needs to be built only once, while the Ball-tree for centroids must be constructed in each iteration of the algorithm as the centroids move. 
% We build a Ball-tree index structure for the data points only once, and the root of the tree is denoted as $R$. Since the clusters' centroids move in every iteration, we build the centroid index in each iteration, with a root node $R_C$.
%\jf{This is confusing -- you first need to explain what these indexes are}

The nodes (denoted as $N$ and $N_{C}$) in the spatial vector and centroid indexes are slightly different; both of them need to store the pivot vector $\mathbf{p}^*$ (the mean of all covered spatial vectors/centroids in the node) and radius $r$ to bound child nodes (or spatial vectors if it is a leaf node with capacity $f$). But each node $N$ of the spatial vector index also stores the number of spatial vectors it covers, denoted as $|N|$, e.g., $|N|=8$ in Fig.~\ref{fig:pruning}.

Furthermore, each index node $N$ (or spatial vector $\mathbf{p}_i$) stores an integer $a(N)$ (or $a(i)$) to denote the id of the cluster it was assigned to in the previous iteration. For the current iteration, we can compute the distance between a spatial vector $\mathbf{p}_i$ and the centroid of its previous cluster $\mathbf{c}_{a(i)}$. If the distance is smaller than the inter bound, i.e., 
\begin{equation}
\|\mathbf{p}_i- \mathbf{c}_{a(i)}\| < \frac{\mathbf{cb}[a(i)]}{2},
\label{equ:interbound1}
\end{equation}
where $\mathbf{cb}[a(i)]$ is defined in Eq. (\ref{equ:inter}), $\mathbf{p}_i$ still belongs to the cluster $a(i)$ in the current iteration \cite{Elkan2003}.

Similarly, if a node $N$ was assigned to the cluster $a(N)$ in the previous iteration, we can compute the distance between $N$ and the centroid of cluster $a(N)$, which denotes $\mathbf{c}_{a(N)}$. If the upper bound on the distance between $N$'s points and $\mathbf{c}_{a(N)}$ is smaller than the inter bound, i.e.,
\begin{equation}
\|N.\mathbf{p}^*- \mathbf{c}_{a(N)}\|+N.r <\frac{\mathbf{cb}[a(N)]}{2}.
\label{equ:interbound}
\end{equation}

Then all the spatial vectors in the node $N$ can be directly assigned to cluster $a(N)$ in the current iteration; otherwise, we search for the two nearest centroids, $\mathbf{c}_{n_1}$ and $\mathbf{c}_{n_2}$, of $N$'s pivot, and denote the corresponding distances as $d_1$ and $d_2$, where $d_2 > d_1$. 
% Otherwise, we search for its two nearest centroids $n_1$ and $n_2$ and the distance $d_1$ and $d_2$ by scanning all $k$ centroids, or with $\mathtt{k}$NN (will describe later).
%which can be accelerated by the centroid index with a complexity of $log_2k$.
If the distance gap $d_2-d_1$ is bigger than $2 N.r$, all the spatial vectors in the node $N$ can be assigned to the cluster with centroid ${\mathbf{c}_{n_1}}$: 
\begin{equation}
d_2 - N.r \ge d_1 + N.r \rightarrow a(N)=n_1.
\end{equation}
%\jf{The references to the red circles are a bit confusing; what about using another feature to distinguish the red circles? Maybe the big is represented with a contiguous line, and the small ones with a doted line?}
%\

If the node $N$ still cannot be assigned, we split $N$ into two (e.g., the two small red dotted circles in Fig.~\ref{fig:pruning}) and repeat the above process for each child node. If the node $N$ is a leaf, we search for the nearest centroid of each spatial vector in $N$ and assign the spatial vectors to their nearest centroids. 

% If the node $N$ is a leaf and we need to assign each of $N$'s points to a cluster, we can set the radius $r$ to zero and change it to search for the nearest centroid. 

The bottleneck in the above process is searching for the nearest centroids of an index node (or a spatial vector). A na\"ive approach would be to compute the distance from the index node (or spatial vector) to each of the centroids, which is computationally expensive. In the following, we use $\mathtt{k}$NN to search for the nearest centroids efficiently. 

\myparagraph{Using $\mathtt{k}$NN to Search for (Two) Nearest Centroids}
To find the two nearest centroids of an index node, 
%
%\jf{suggest: checking if all $k$ centroids are not pruned by the inter bound has a worst...}
checking if all $k$ centroids are not pruned by the inter bound has a worst-case time complexity of $O(k)$. 
Here, we use the $\mathtt{k}$NN search method based on the index structure of the centroids. This method reduces the time complexity to $O(\log_2 k)$ on average, by pruning a set of centroids in a centroid index node if its lower bound to the query vector $\mathbf{q}$ is larger than the current results held in a priority queue $H$.
Initially, $H$ is filled with arbitrarily large numbers if no result has been found.
We can prune certain centroid nodes in advance by deriving a tight upper bound on the distance from the query vector $\mathbf{q}$ to its two nearest centroids, as detailed below.
% if we can estimate the upper bound distance to the nearest centroid.

%kNN bound用来预测一个error bound,跟下列的kNN bound对比，如果合理直接找一个bound出来，来进一步紧缩这个bound范围，减少kNN查询范围。
\myparagraph{$\mathtt{k}$NN Bounds Inherited from Parent Nodes}
%To accelerate the $\mathtt{k}$NN search, we can be accelerated with pruning nodes with a lower bound distance bigger than the current nearest one.
To further prune centroid nodes during the $\mathtt{k}$NN search, we compute an upper bound on the distance from the pivot of a node ($N^{'}.\mathbf{p}^*$) to its two nearest neighbors:
\begin{equation}
\begin{split}
ub_1(N^{'}.\mathbf{p}^*) = d_1(N.\mathbf{p}^*) + N.r,\\
ub_2(N^{'}.\mathbf{p}^*) = d_2(N.\mathbf{p}^*) + N.r,
\end{split}
\end{equation}
where $N$ is the parent node of $N^{'}$; $d_1(N.\mathbf{p}^*)$ and $d_2(N.\mathbf{p}^*)$ are the distances from $N.\mathbf{p}^*$ to its two nearest centroids, as shown in Fig.~\ref{fig:pruning}.  
% inherited from the parent node which already searched the nearest two neighbors, as shown in Figure~\ref{fig:pruning}(b),
% \begin{equation}
% \begin{split}
% ub_1(N^{'}.p) = d_1(N.p) + N.r,\\
% ub_2(N^{'}.p) = d_2(N.p) + N.r,
% \end{split}
% \end{equation}
%After we compute the distance to its nearest centroid, we can get another bound using the inter bound based on triangle inequality:
%\begin{equation}
%ub(N^{'}.p) = \min\big(ub(N^{'}.p), \|N.p- c_{a(i)}\|+cb[a(i)]\big)
%\end{equation}
When searching for the two nearest centroids of $N^{'}.\mathbf{p}^*$, we can prune a centroid node $N_{C}$, if the lower bound on the distance between the centroids in $N_{C}$ and $N^{'}.\mathbf{p}^*$ is larger than ${ub}_2(N^{'}.\mathbf{p}^*)$,
% With the second nearest centroid distance's upper bound $ub_2(N^{'}.p)$, when we traverse the centroid index with the pivot point $p$ of point node $N$, we can prune $N_{C}$ if a centroid node $N_{C}$ has a lower bound bigger than it:
\begin{equation}
\|N^{'}.\mathbf{p}^*-N_{C}.\mathbf{p}^*\|-N_{C}.r > {ub}_2(N^{'}.\mathbf{p}^*).
\end{equation}

If we search for the nearest centroid of $N^{'}.\mathbf{p}^*$, we can replace ${ub}_2(N^{'}.\mathbf{p}^*)$ with ${ub}_1(N^{'}.\mathbf{p}^*)$ in the above inequality for pruning centroid nodes.

\begin{algorithm}[]
\small
\label{alg:kmeans}
    \caption{\texttt{Accelerator}($k$, $\mathbf{D}$, $M$)}
    \KwIn{$k$: the number of clusters, $\mathbf{D}$: dataset, $M$: available main memory}
    \KwOut{$k$ centroids: $\mathbf{C}=\{c_1,\dots, c_k\}$}
    %\rule{7.2cm}{0.4pt}
	
    Create Ball-tree on $\mathbf{D}$ according to $M$, get the root node $R$\;

    %$\mathbf{C}\leftarrow$ \texttt{{InitializeCentroids}}($k, \mathbf{D}$)\;
    Initialize centroids $\mathbf{C}$ according to $k$ and $\mathbf{D}$\;

    $it \leftarrow 1$\;
    
    \While{did not converge}{
        Create Ball-tree on $\mathbf{C}$ and get root node $R_C$\;
    	\ForEach{$\mathbf{c}_j \in \mathbf{C}$}{
            Set $ub$ as $\infty$ if $it=1$ else set $ub$ using Eq. (\ref{equ: ub})\;
    		$[Q, H] \leftarrow {\mathtt{k}\texttt{\textbf{NN}}}(2, \mathbf{c}_j, R_C, ub)$\;
      	$\mathbf{cb}[j] \leftarrow H[2]$; \tcp{defined in Eq.(\ref{equ:inter})}
    	}
		
        $[S, \mathbf{sv}]\leftarrow$\texttt{\textbf{Assign}}($R$, $R_C$, $\infty$)\;

    	\ForEach{$\mathbf{c}_j \in \mathbf{C}$}{
        \vspace{-0.25em}
    		Refine centroid: $\mathbf{c}_j\leftarrow \frac{\mathbf{sv}(j)}{|S_j|}$, and compute $\Delta[j]$ \label{line:refine};
    		\vspace{-0.25em}
    	}
        $it \leftarrow it + 1$\;
    }
	\KwRet $\mathbf{C}$;
	\noindent\rule{8cm}{0.4pt}
	
%%%%%%%% function: Assign %%%%%%%%
    \SetKwFunction{FMain}{\textbf{Assign}}
    \SetKwProg{Fn}{Function}{:}{}
    \label{alg:assign}
    \Fn{\FMain{$N$, $R_C$, $ub$}}{
    \KwIn{$N$: node (or spatial vector) to be assigned, $R_C$: root node of centroid index, $ub$: upper bound}
    \KwOut{$S$: cluster with nodes \& spatial vectors, $\mathbf{sv}$: sum vector of cluster $S$}
	\vspace{-0.7em}
	\rule{7.2cm}{0.4pt}
		
    \eIf{$N$ is node}{
    \vspace{-0.25em}
        \If{$\|N.\mathbf{p}^*- \mathbf{c}_{a(N)}\| + N.r < \frac{\mathbf{cb}[a(N)]}{2}$}{
        	Assign node $N$ to cluster $a(N)$\;%\tcp*{\small maintain} 
            \KwRet
        }
        
	$[Q, H] \leftarrow {\mathtt{k}\textbf{NN}}(2, N.\mathbf{p}^*, R_C, ub)$\;
    $\mathbf{c}_{n_1}, \mathbf{c}_{n_2}, d_1, d_2 \leftarrow Q[1], Q[2], H[1], H[2]$\;
   
	\eIf{$(d_2 - d_1) > 2*N.r$}{
		\If{$a(N) \neq n_1$}{Update $S_{a(N)}$, $\mathbf{sv}(a(N))$ and $S_{n_1}$, $\mathbf{sv}(n_1)$\;\label{line:move}}
				
		Assign node $N$ to cluster $n_1$\;
            \KwRet
		\vspace{-0.5em}
	}{
		\vspace{-0.5em}
		\ForEach{child node or spatial vector $N^{'}$ of $N$}{
                \vspace{-0.25em}
			\textbf{Assign}($N^{'}$, $R_C$, $d_2+N.r$);
		}
	}
	\vspace{-0.35em}
    }{
	\vspace{-0.35em}
         \If{$\|N - \mathbf{c}_{a(N)}\| < \frac{\mathbf{cb}[a(N)]}{2}$}{
    	Assign spatial vector $N$ to cluster $a(N)$\; 
            \KwRet
        }
        
	$[Q, H] \leftarrow {\mathtt{k}\textbf{NN}}(1, N, R_C, ub)$\;
    $\mathbf{c}_{n_1} \leftarrow Q[1]$\;

	\If{$a(N) \neq {n_1}$}{Update $S_{a(N)}$, $\mathbf{sv}(a(N))$, $S_{n_1}$, and $\mathbf{sv}(n_1)$;\label{line:move1}}
			
	Assign spatial vector $N$ to cluster $n_1$;
    }

    \KwRet [$S$, $\mathbf{sv}$];
    }
	
	%\vspace{-0.5em}
	\noindent\rule{8cm}{0.4pt}
%	\vspace{0.5em}
% Pseudocode for Ball-tree knn query
\SetKwFunction{FMain}{$\mathtt{k}$\textbf{NN}}
\SetKwProg{Fn}{Function}{:}{}
\label{alg:knn}
\Fn{\FMain{$\mathtt{k}$, $\mathbf{q}$, $N_{C}$, $ub$}}{
\KwIn{$\mathtt{k}$: the number of neighbors, $\mathbf{q}$: query vector, $N_{C}$: centroid node, $ub$: upper bound distance to the nearest centroid}
\KwOut{$Q$: a priority queue holding $\mathtt{k}$NN, $H$: a priority queue holding distances of the $\mathtt{k}$NN}
\vspace{-0.7em}
\rule{7.2cm}{0.4pt}\\

    Initialize the distances in the priority queue $H$ to $ub$\;
    \eIf{$N_{C}$ is a leaf node}{
        \ForEach{spatial vector $\mathbf{p}_i \in N_{C}$}{
            $d \leftarrow \|\mathbf{p}_i - \mathbf{q}\|$\;
    	%Compute $\|\mathbf{p}_i - \mathbf{q}\|$\, and update result heap $H$;
    				%	$t\leftarrow$ $\mathtt{k}$-th result's distance in $H$\;
    				
    	\If{$d < H[k]$}{
                Update $Q$ and $H$ using $\mathbf{p}_i$ and $d$\;
            }
        }
        \vspace{-0.35em}
    }{
        \vspace{-0.35em}
        \ForEach{child node $N_{C}^{'}$ of $N_{C}$}{
    	$d \leftarrow \|\mathbf{q}-N_{C}^{'}.\mathbf{p}^*\|-N_{C}^{'}.r$\;
    	\If{$d < H[k]$}{
                \textbf{$\mathtt{k}$NN}($\mathtt{k}$, $\mathbf{q}$, $N_{C}^{'}$, $H[k]$);
            }
        }
    }
    \KwRet $[Q, H]$;
}
\end{algorithm}
%\setstretch{1}

\myparagraph{Accelerating Inter Bound Computation}
To compute a tight inter bound for a centroid $\mathbf{c}_j$, we need to find the minimum distance from $\mathbf{c}_j$ to other centroids. In contrast to \cite{Xia2020} which computes the pairwise distances between centroids, we use $\mathtt{k}$NN to search for the nearest centroid of $\mathbf{c}_j$ efficiently. We also derive an upper bound on the distance from $\mathbf{c}_j$ to its nearest centroid to further prune centroid nodes. Let $\mathbf{cb}[j]$ denote the distance from $\mathbf{c}_j$ to its nearest centroid in the previous iteration of the algorithm; $\Delta[j]$ denote the drift of $\mathbf{c}_j$; and $\max(\Delta)$ denote the maximum drift of $\mathbf{c}_j$'s nearest centroid, the upper bound is defined as:
\begin{equation} \label{equ: ub}
{ub}=\mathbf{cb}[j]+\Delta[j]+ \max(\Delta).
\end{equation}

In summary, we have used $\mathtt{k}$NN to accelerate various components of our approach: 1) inter bound computation with $\mathtt{k}=2$; 2) node assignment with $\mathtt{k}=2$; and 3) spatial vector assignment with $\mathtt{k}=1$.
All of them can be accelerated by an update-free upper bound from parent nodes.
% With the distance $cb[j]$ to the previous nearest centroid, we further use its drift $\Delta[j]$ and the maximum bound drift $max(\Delta)$ of its nearest centroid to estimate the upper bound distance for each centroid $c_j$, which can further achieve pruning on the centroid index.
% \begin{equation}
% ub=cb[j]+\Delta[j]+max(\Delta),
% \end{equation}

% To find the nearest centroid and get the inter bound for each centroid,
% we do not compute all-pair distances like \cite{Xia2020}. 
% Instead, we conduct the nearest centroid search.
% To achieve this, we set $k=2$ to search as the nearest one is the centroid itself.
% With the distance $cb[j]$ to the previous nearest centroid, we further use its drift $\Delta[j]$ and the maximum bound drift $max(\Delta)$ of its nearest centroid to estimate the upper bound distance for each centroid $c_j$, which can further achieve pruning on the centroid index.
% \begin{equation}
% ub=cb[j]+\Delta[j]+max(\Delta),
% \end{equation}
%For the inner bound which needs to find the nearest neighbor for every centroid, we use the built index to conduct a self-join \cite{Jacox2007a} operation, and save much time.

%\sheng{To dos: \myparagraph{Accelerating \kmeans++} In the initialization, every point also need to find the nearest neighbor centroid to assign. We further utilize the centroid index to accelerate, we can conduct an individual part to test.}

\subsection{Algorithm Design}
%\jf{Should the title be pick-means Algorithm?}
%\jf{Since you are describing how the algo works, you should use "it" to refer to the algorithm, and not "we" the authors}
%
Algorithm~\ref{alg:kmeans} {shows the process of pruning mechanism} over \pick.
After creating the spatial vector index on $\mathbf{D}$ and the centroid index on the initial $k$ centroids, \pick uses recursion to traverse the spatial vector index and centroid index to conduct the assignment with a bound-armed $\mathtt{k}$NN search.
After assigning all the spatial vectors to their nearest centroid, it refines the centroids and checks whether any of the centroids move; if so, it continues.
To refine the new centroid efficiently, \pick maintains a dynamic sum vector $\mathbf{sv}(j)$ for each cluster with a unique id $j$.
It updates $\mathbf{sv}(j)$ when a spatial vector $\mathbf{p}$ moves in ($\mathbf{sv}(j) = \mathbf{sv}{(j)} + \mathbf{p}$) or out ($\mathbf{sv}(j) = \mathbf{sv}{(j)} -\mathbf{p}$) (see Lines~\ref{line:move} and \ref{line:move1}), where $\mathbf{p}$ can be replaced by $N.\mathbf{p}^* \cdot |N|$ if a node $N$ moves.
Finally, a new centroid $\mathbf{c}_j$ can be computed by $\frac{\mathbf{sv}(j)}{|S_j|}$ in Line~\ref{line:refine}.
%Without complex bound computation and maintenance, 
Our algorithm can be easily implemented with two recursive traversal functions presented below. Notably, We analyze the time complexity of the proposed pruning
mechanism in Appendixes~\ref{sec:Complexity Analysis} due to the page limitation.

\myparagraph{Recursive Traversal on Spatial Vector Index} 
The function \textbf{Assign} traverses the spatial vector index to assign spatial vectors in batch or one by one. From the root node of the spatial vector index, the function searches for the two nearest centroids using \textbf{$\mathtt{k}$NN}.
After computing the distance gap $d_2-d_1$, it checks whether the centroid can be pruned; if not, it sends the bound to its child nodes and performs another \textbf{Assign} operation recursively. 
%\jf{suggest to add: recursively}.

\myparagraph{Recursive Traversal on Centroid Index} 
Function $\mathtt{k}$\textbf{NN} recursively searches the centroid index to get one or two nearest centroids, using an upper bound ${ub}$ to prune certain centroid nodes, and ${ub}$ is initialized as the bound from the parent node and is updated with the latest centroid's distance found in $H$. A centroid node can be pruned if the lower bound on the distance from the query vector to each of the centroids in the node is greater than ${ub}$.

%devices with limited-resource
\section{Lightweight Cost Estimator}
\label{sec:costmodel}
\begin{figure}
	\centering
  \includegraphics[width=.478\textwidth]{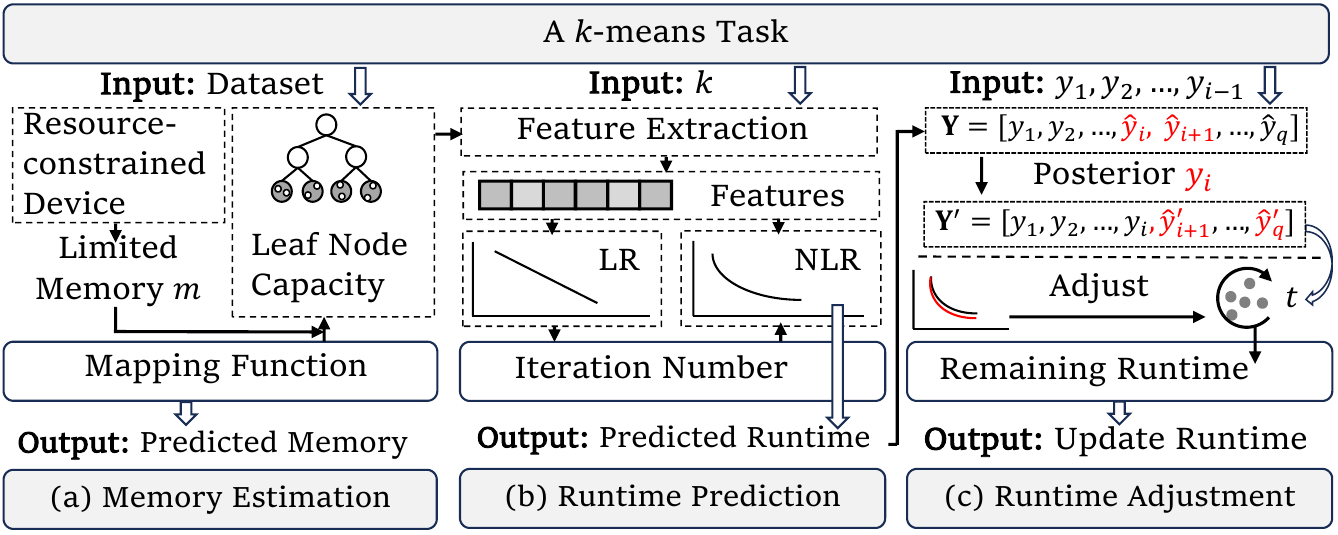}
	\vspace{-1em}
	\caption{Overview of our lightweight cost estimator, where \( y_i \) denotes the actual runtime for the \( i \)-th iteration (\( i = 1, 2, \dots, q \)), and \( \hat{y}_j \) represents the predicted runtime for the \( j \)-th iteration (\( j = 1, 2, \dots, q \)).}
	\label{fig:overview}
	\vspace{-2em}
\end{figure}

\myparagraph{Overview} 
We design a lightweight cost estimator to accelerate the \kmeans algorithm. Firstly, as shown in Fig.~\ref{fig:overview}(a), we propose a memory estimation method to predict memory costs by building a mapping function between the index and the memory. This method also allows us to create memory-tunable indexes under memory constraints, thereby accelerating the \( k \)-means tasks. Secondly, as shown in Fig.~\ref{fig:overview}(b), we predict the runtime of the $k$-means task by estimating the iteration number using a \textit{linear regressor} (\textbf{LR}) and the runtime of each iteration using a \textit{non-linear regressor} (\textbf{NLR}). Finally, as shown in Fig.~\ref{fig:overview}(c), we monitor the progress of the $k$-means task by dynamically
updating the remaining runtime. Specifically, we use posterior information from the last iteration of the $k$-means task to adjust the predicted runtime using a \textit{Gaussian Process} (\textbf{GP}) with an asymmetric kernel function.

\subsection{Memory Cost Estimation}
\label{sec:memory}
The memory required for the \kmeans algorithm includes storing the dataset, maintaining the bounds, and the memory occupied by the indexing structure. 
Besides storing the dataset, the additional memory required is solely related to the indexing structure due to the fact that \pick does not maintain any bound. 
Hence, we estimate the memory cost of the index by establishing a mapping function between the leaf node capacity and the memory cost, denoted as $m$.
For the index (using the balanced Ball-tree structure), each node includes a vector (a center of each partitioned sub-space, 3 dimensions), three floats (radius \( r \), number of spatial vectors covered, cluster ID), and two pointers to child nodes (left and right) or a set of spatial vectors in leaf nodes (up to capacity \( f \)). 
Thus, we estimate the memory cost of a leaf node as $3+3+f$, and an internal node as $3+3+2=8$.
% Moreover, we can determine that the index has a total of $2 \lceil \frac{n}{f} \rceil - 1$ nodes, among which $\lceil \frac{n}{f} \rceil$ are leaf nodes.
Then the overall memory cost (number of floats) of all the nodes is:
	\begin{equation}
	\begin{split}
	\mathcal{M}(n, f)
&= \lceil \frac{2n}{f} \rceil \cdot(6+f)+(\lceil \frac{2n}{f} \rceil - 1)\cdot8\\ 
&\approx 2n+\frac{28n}{f}-16,
%&= \frac{2n}{f}\cdot(6+f)+\frac{2n}{f}(1-2^{1-\log_2\frac{2n}{f}})\cdot8\\ 
%	&=2n+\frac{12n}{f}+\frac{16n}{f}\cdot(1-2^{1-log_2\frac{2n}{f}})\\
%	&=2n+\frac{28n}{f}-\frac{32n\cdot2^{-\log_2\frac{2n}{f}}}{f}\\
%	&=2n+\frac{28n}{f}-16
	\end{split}
	\end{equation}
	% where $\frac{2n}{f}$ and $\frac{2n}{f}(1-2^{1-\log_2\frac{2n}{f}})$ are the numbers of leaf nodes and internal nodes}.
where $\lceil \frac{2n}{f} \rceil$ and $\lceil \frac{2n}{f} \rceil - 1$ are the numbers of leaf nodes and internal nodes, respectively. This estimation is based on the assumption that each leaf node has $\frac{f}{2}$ spatial vectors, and the balanced Ball-tree with a height $\lceil \log_2\frac{2n}{f} \rceil$. It is worth noting that, in the real case, most nodes are not fully filled -- they are half full on average. Therefore, we double the number of leaf nodes and internal nodes. Similarly, the centroid index also occupies $\mathcal{M}(k, f)$ units of memory.\footnote{Here, we assume using a 64bit system on resource-constrained devices.}

Moreover, the array used to indicate which cluster each spatial vector is assigned to will occupy \( n \) integers. This array stores the cluster IDs for the spatial vectors and helps identify which clusters they belong to.
Hence, compared to Lloyd's algorithm, \pick requires additional memory, which can be described as follows:
%Each cluster will maintain a list to store the assigned node and ids, if we want to accelerate the refinements without scanning all the points based on the assigned cluster array. 
\begin{equation}\small
\label{eqn:memory_estimation}
%m = \mathcal{M}(n, f) + \mathcal{M}(k, f) + n  \approx 2n+\frac{28n}{f}-16 + 2k+\frac{28k}{f}-16+n. 
m = \mathcal{M}(n, f) + \mathcal{M}(k, f) + n  \approx (2+\frac{28}{f})(n+k)-32+n. 
\end{equation}

Based on this analysis, we can adjust the node capacity $f$ according to the available memory $m$ when clustering must be performed in resource-constrained devices,
\iffalse
\myparagraph{{Memory-aware Parameterization}}
According to the available memory, we can set the number of nodes and use inter bounds (with assigned clusters) to prune.
We show the priority: 1) centroid index (small), 2) spatial vector index with various node sizes, 3) inter bounds, and 4) assigned cluster id.
All the bounds can be controlled and configured according to the available memory.
%We do not need to store any bounds for any node or point across iterations. 
We need to record which cluster the node or spatial vector is assigned to, which can save a lot of space.

%If we do not have space for the to-parent distance, we will return $\infty$ every time when we call $N.toParentDis()$.
\fi

\myparagraph{Memory-tunable Index}
A common index structure, such as the kd-tree and cover-tree used in \cite{Curtin2017}, needs to store the leaf nodes as two spatial vectors at most, and the memory cost is at least $\mathcal{M}(n, 2)$, which is much higher than our index, which utilizes the ball-true structure.
%A common index structure, such as the kd-tree and cover-tree used in \cite{Curtin2017}, 
Instead, we automatically configure the leaf node capacity $f$ (the leaf node size of two index trees) based on the memory constraint, denoted as $m^{'}$. Specifically, under a given memory constraint $m^{'}$, $f$ can be calculated using Eq.~\eqref{eqn:memory_estimation} as follows:
\begin{equation}
\label{eqn:aware_memory}
f \approx \frac{28(n+k)}{m^{'}-3n+32-2k}. 
\end{equation}
%In Table~\ref{tab:memory}, we compare our algorithm with other algorithms in terms of tunable memory cost. In the experimental evaluation, we present a detailed comparison with the two most recent work \cite{Curtin2017} and \cite{Xia2020}.

Overall, no bound is maintained for each spatial vector, and although we need to maintain an inter bound for each centroid in each iteration, the cost is negligible as $k \ll n$. Hence, the size of our index can be auto-configured according to the available memory via tuning the leaf node capacity $f$.
%\jf{You mention "only two bounds", but then you mention only one -- the inter bound}

\subsection{Runtime Prediction}
\subsubsection{Non-Linear Regressor}
\label{sec:regressor}
Unlike traditional methods \cite{KapusC19,TueroB21} that rely on training samples to directly predict $k$-means task runtime, denoted as $t$, our approach estimates the total runtime by predicting the iteration number and each iteration's runtime, respectively. Firstly, we estimate the $k$-means iteration number by using a linear regressor. Specifically, instead of using a positive integer to represent the iteration number, denoted as \(\upsilon\), we use a dummy array, denoted as $\mathbf{u}$, which is composed of 1s in the first \(\upsilon\) positions and 0s in the remaining positions. For example, if $\upsilon = 2$ and the maximum iteration number, denoted as $q$, is 5, then $\mathbf{u} = [1, 1, 0, 0, 0]$. 

Then, we predict each iteration's runtime by designing a polynomial expression in a non-linear regressor. Finally, we calculate the total runtime as shown below:
\begin{equation}
    t = \sum^q_{i=1} u_i \times \hat{y}_i,
\end{equation}
where $u_i$ is the value at the $i$-th position in \(\mathbf{u}\), and $y_i$ is the predicted runtime of the $i$-th iteration.

\myparagraph{Iteration Number Estimation} 
We predict the iteration number $\upsilon$ using a linear regressor, such as multiple linear regression \cite{su2012linear}, which builds the function from the meta-feature to $\upsilon$. Notably, extracting a meta-feature to describe (or represent) a dataset, such as $n$, $k$, and $d$, is not informative. Hence, in addition to these features, we also extract novel and more complex features to capture certain properties of data distribution based on our index. Specifically, the index construction actually conducts a more in-depth scan of the spatial vectors and reveals whether the spatial vectors assemble well in the space. Hence, the information can include tree depth, number of leaf nodes, number of internal nodes, and average spatial vectors per leaf node.

\myparagraph{Building Non-linear Regressor}
We design a non-linear regressor with \(\mathbf{u}\) to model how meta-features, including \(n\), \(k\), \(d\), and \(f\), affect the runtime of $k$-means. We notice that the extracted meta-features are not independent. For example, $n$ and $f$ jointly determine the index structure, which affects the efficiency of $\mathtt{k}$NN and affects the runtime of the assignment process in $k$-means. Therefore, we need to consider interaction terms (or \textit{interaction feature}), such as $nf$. The regressor considering interaction feature can be expressed using a polynomial expression as follows:
\begin{equation}
    \hat{y}_j = \sum_{i_1, i_2, \ldots, i_{\lambda} = 0}^\lambda u_j\beta_{i_1 i_2 \ldots i_\lambda} x_{j1}^{i_1} x_{j2}^{i_2} \ldots x_{j\lambda}^{i_\lambda}+e,
\end{equation}
where $\lambda$ is the number of meta-features, $\beta_{i_1 i_2 \ldots i_\lambda}$ is the regression coefficient, ($x_{j1}$, $x_{j2}$, $\cdots$, $x_{j\lambda}$) are the meta-features obtained for the $j$-th iteration, and $e$ is the residual term. Then the runtime of the $k$-means task, denoted as $\hat{y}$, can be represented as follows:
\begin{equation}\tiny
\label{eqn:regression_expression}
\hat{y} = \begin{bmatrix} 1 \\ 1 \\ \vdots\\1 \end{bmatrix}'
\begin{bmatrix} u_{1} & 0 & \cdots & 0\\ 0 & u_{2} & \cdots & 0 \\ \vdots & 0 & \cdots & \vdots  \\ 0 & 0 & \cdots & u_{q} \end{bmatrix} \renewcommand{\arraystretch}{1.5}\begin{bmatrix} x_{11} & x_{12} & \cdots & \sum^\lambda_{i=1} x^\lambda_{1i}\\   x_{21} & x_{22} & \cdots &  \sum^\lambda_{i=1} x^\lambda_{2i}\\ \vdots & \vdots & \cdots & \vdots  \\ x_{q1} & x_{q2} & \cdots &  \sum^\lambda_{i=1} x^\lambda_{qi} \end{bmatrix}\begin{bmatrix} \beta_{1} \\ \beta_{2} \\ \vdots\\\beta_{\sum^\lambda_{i=1}\binom{\lambda}{i}} \end{bmatrix}+e.
\end{equation}

For simplicity, we represent Eq.~\eqref{eqn:regression_expression} with the following equation:
\begin{equation}
    \hat{y}= \mathbf{1}' \mathbf{u} \mathbf{x}\mathbf{b}+e,
\end{equation}
where $\mathbf{1} \in \mathbb{R}^{1 \times q}$, $\mathbf{u} \in \mathbb{R}^{q \times q}$, $\mathbf{x} \in \mathbb{R}^{q\times \sum^\lambda_{i=1}\binom{\lambda}{i}}$, and $\mathbf{b} \in \mathbb{R}^{\sum^\lambda_{i=1}\binom{\lambda}{i} \times 1}$. Given \( n_1 \) samples, we represent \( [\mathbf{1}' \mathbf{u} \mathbf{x}_1, \mathbf{1}' \mathbf{u} \mathbf{x}_2, \dots, \mathbf{1}' \mathbf{u} \mathbf{x}_n]' \) as \( \mathbf{X} \). The resulting non-linear model then can be solved using ordinary least squares (OLS) \cite{maulud2020review}. The solution for \( \mathbf{b} \) is as follows:
\begin{equation}
\label{eqn:optimization}
\mathbf{b} = [\sum^n_{i=1}\mathbf{1}' \mathbf{u} \mathbf{x}_i\mathbf{1}' \mathbf{u} \mathbf{x}_i]^{-1}[\mathbf{1}' \mathbf{u} \mathbf{x}_1,\mathbf{1}' \mathbf{u} \mathbf{x}_2,\cdots,\mathbf{1}' \mathbf{u} \mathbf{x}_n]\mathbf{y}.
\end{equation} 

Therefore, we feed \( \mathbf{u} \) and \( \mathbf{x} \) into Eq.~\eqref{eqn:optimization} to obtain \( \mathbf{b} \), and subsequently use the trained regressor to predict \( t \).

\subsubsection{Runtime Adjustment with GP}
\label{sec:GP}
We design a GP with an asymmetric kernel function to iteratively adjust the predicted runtime $\mathbf{\hat{Y}}$, hence monitoring the progress of $k$-means.
Specifically, once the actual runtime $y_i$ for the $i$-th iteration becomes available at the end of that iteration, we can figure out the posterior information by examining the difference between \( y_i \) and \( \hat{y}_i \) to refine $\mathbf{\hat{Y}}$. This way, with each completed iteration, we can further adjust the estimated runtime.

A commonly used method for adjusting the predicted runtime of iterative algorithms is \textit{Weighted Average} \cite{Mitchell2022}. This method assumes that the predicted time for the next iteration depends solely on previous iterations. However, in practice, information from the current iteration can affect the runtime of all subsequent iterations. For example, if $k$-means converges within the current iteration, the runtime for all future iterations will be 0, as the $k$-means task is complete. To address this limitation, GP is a better choice because GP adjusts the prediction of runtime for all iterations based on the degree of correlation between subsequent and current iterations.

\myparagraph{Formulation of GP}We build a GP over the predicted runtime, which can be expressed as follows:
\begin{equation}
\label{eqn: GP definition}
g(i) \sim \text{GP}(\mu(i), cov(i, i^{'})),
\end{equation}
where \( g(i) \) is the ratio between the predicted runtime \( \hat{y}_i \) and the actual runtime \( y_i \) for the \( i \)-th iteration, $\mu(i)$ represents the mean of $g(i)$, and $cov(i, i^{'})$ represents the \textit{kernel function} (or covariance function) that describes the correlation between the $i$-th iteration and the $i^{'}$-th iteration. Notably, when the given $k$-means task has not yet run, we assume perfectly accurate predictions, i.e., \( \hat{y}_i = y_i \), which implies \( g(i) = 1 \). Under this condition, the initial mean function of the GP becomes a constant function equal to 1 for all iterations.

%\begin{figure*}
%	\centering
%\includegraphics[width=1\textwidth]{K-means-iteration.pdf}
%	\vspace{-2em}
%	\caption{The distribution of runtime for each iteration of \kmeans algorithms.}
%	\label{fig:boxplot}
%    \vspace{-1.5em}
%\end{figure*}

\myparagraph{Asymmetric Kernel Function of GP}Unlike a classical GP \cite{JayasumanaHSLH15}, where posterior information can be bidirectional. For example, A commonly used kernel function is the Radial Basis Function (KBF) kernel \cite{JayasumanaHSLH15}, which can be shown as follows:
% Such as the RBF kernel defined as follows:
\begin{equation}
\label{eqn: RBF kernel}
cov(i, i^{'}) = \exp\left(-\frac{\|i^{'} - i\|^2}{2\sigma^2}\right),
\end{equation}
where $\sigma$ is a hyperparameter for adjusting the correlation between the $i$-th iteration and the $i^{'}$-th iteration. Here we need to account for the fact that posterior information from the current iteration of $k$-means affects only subsequent iterations (i.e., completed iterations influence upcoming ones), which means the correlation should only propagate in the direction of increasing $i$. Therefore, we design the specific expressions for $\text{cov}(i, i^{'}) $. To simulate the unidirectional propagation of correlation in an iterative process, we design a new kernel function, which is shown as follows:
\begin{equation}
\label{eqn: kernel function}
cov(i, i^{'}) = 
\begin{cases} 
0, & \text{if } i^{'}-i \leq -1; \\
\exp\left(-\frac{h(i^{'} - i)^2}{2\sigma^2}\right), & \text{if } i^{'}-i > -1;
\end{cases}
\end{equation}
where the iteration numbers $i^{'}$ correlated with $i$ are restricted to the interval $(i-1, +\infty)$. This implies that the actual runtime of the $i$-th iteration only affects the iterations within the range of $(i-1, +\infty)$. Moreover, To ensure that the convergence function is continuously differentiable over its domain, we design \( h(\delta) \) as follows:
\begin{equation}
\label{eqn: mapping g(i)}
h(\delta) = 
\begin{cases} 
\ln{(\delta + 1)}, & \text{if } -1 < \delta \leq 0; \\
\delta, & \text{if } \delta > 0;
\end{cases}
\end{equation}
where,  $h(\delta)$ ensures differentiability of $cov(i,i^{'})$ at \((i^{'}-i) = -1\), thus guaranteeing the differentiability of the kernel function in its' domain.

\section{Experiments}
\label{sec:exp}

\begin{table}[]
\ra{1.2}
\caption{An overview of the datasets (M for million).}
\label{tab:datasets}
\vspace{-1em}
\begin{tabular}{cccc}
\toprule
\textbf{Dataset}                  & \textbf{Dimensionality} & \textbf{Scale} & \textbf{Description}              \\ \midrule
\Tdrive   & 2              & 1M             & Trajectory data point    \\
\Porto    & 2              & 1M             & Trajectory data point    \\
\ArgoAVL  & 2              & 1M            & Trajectory data point    \\
\ArgoPC   & 3              & 1M             & Point cloud data         \\
\DRD      & 3              & 0.43M             & Point cloud data         \\
\Shapenet & 3              & 1M             & Point cloud data         \\
\ApollTD  & 128              & 0.5M             & Embedded trajectory data \\
\ArgoETD & 256              & 0.5M             & Embedded trajectory data    \\ \bottomrule
\end{tabular}
\vspace{-2.5em}
\end{table}

% \usepackage{multirow}
\begin{table*}[]
\caption{The performance of \pick in terms of runtime.}
\label{tab:runtime}
\vspace{-1em}
%\ra{1.1}
\setlength{\tabcolsep}{4.8pt}
\centering
\begin{tabular}{cccccccccccc}
\toprule
                   \textbf{Dataset}                    &   Settings   & \texttt{Lloyd}        & \texttt{NoBound}       & \texttt{Dual-tree}      & \texttt{Hamerly}       & \texttt{Drake}      & \texttt{Yinyang}        & \texttt{Elkan}              & \texttt{NoInB}       & \texttt{No$\mathtt{k}$NN}        & \pick               \\ \midrule
\multirow{3}{*}{\Tdrive}                     & $k=10^2$ & 128.10   & 954.26  & 65.47  & 34.31   & 88.21  & 70.25    & 21.52          & 19.55   & 30.16    & \textbf{13.13}  \\
                                       & $k=10^3$ & 1234.78  & 385.21  & 98.87  & 295.60  & 541.52 & 649.61   & 159.23         & 385.21  & 285.01   & \textbf{28.49}  \\
                                       &$k=10^4$ & 24755.76 & 6225.76 & 601.22 & 5853.36 & N/A    & 13954.17 & N/A            & 6225.76 & 15547.69 & \textbf{211.36} \\ \midrule
\multirow{3}{*}{\Porto}                     & $k=10^2$ & 131.11   & 1119.78 & 72.51  & 36.71   & 72.95  & 71.22    & 23.61          & 23.05   & 32.61    & \textbf{15.13}  \\
                                       & $k=10^3$ & 1227.00  & 412.38  & 102.86 & 298.38  & 520.28 & 642.69   & 162.86         & 412.38  & 314.19   & \textbf{32.07}  \\
                                       & $k=10^4$ & 12295.26 & 3036.00 & 300.22 & 2950.58 & N/A    & 6933.70  & N/A            & 3036.00 & 8822.68  & \textbf{237.80} \\ \midrule
\multirow{3}{*}{\ArgoAVL}                     & $k=10^2$ & 133.59   & 140.65  & 61.95  & 34.15   & 84.61  & 68.37    & 19.58          & 19.46   & 31.80    & \textbf{10.21}  \\
                                       & $k=10^3$ & 1261.24  & 316.66  & 101.94 & 296.20  & 757.24 & 639.45   & 160.00         & 316.66  & 378.36   & \textbf{25.93}  \\
                                       & $k=10^4$ & 12512.56 & 3093.48 & 285.84 & 2886.71 & N/A    & 6853.31  & N/A            & 3093.48 & 8858.64  & \textbf{103.83} \\ \midrule
\multicolumn{1}{l}{\multirow{3}{*}{\ArgoPC}} & $k=10^2$ & 135.33   & 112.83  & 43.82  & 42.76   & 77.42  & 76.00    & 19.22          & 11.68   & 13.69    & \textbf{8.17}   \\
\multicolumn{1}{l}{}                   & $k=10^3$ & 1319.38  & 301.65  & 63.82  & 387.85  & 542.36 & 711.50   & 161.41         & 301.65  & 238.85   & \textbf{16.85}  \\
\multicolumn{1}{l}{}                   & $k=10^4$ & 13247.99 & 3334.37 & 270.62 & 3824.05 & N/A    & 7399.46  & N/A            & 3334.37 & 9316.83  & \textbf{78.56}  \\ \midrule
\multicolumn{1}{l}{\multirow{3}{*}{\DRD}} & $k=10^2$ & 59.16    & 34.50   & 19.37  & 18.98   & 19.25  & 33.29    & 9.25           & 21.81   & 28.03    & \textbf{6.71}   \\
\multicolumn{1}{l}{}                   & $k=10^3$ & 573.37   & 135.14  & 37.52  & 165.94  & 230.50 & 309.79   & 75.27          & 135.14  & 430.44   & \textbf{21.99}  \\
\multicolumn{1}{l}{}                   & $k=10^4$ & 5754.00  & 1724.97 & 188.54 & 1631.71 & 2546.64   & 3255.45  & 853.89           & 1724.97 & 5545.52  & \textbf{47.96}  \\ \midrule
\multicolumn{1}{l}{\multirow{3}{*}{\Shapenet}} &$k=10^2$& 139.69   & 143.99  & 48.89  & 44.80   & 49.88  & 80.66    & \textbf{24.05} & 87.02   & 260.27   & 33.54           \\
\multicolumn{1}{l}{}                   & $k=10^3$ & 1352.89  & 319.90  & 58.82  & 386.38  & 557.32 & 741.57   & 178.24         & 319.90  & 1174.97  & \textbf{77.57}  \\
\multicolumn{1}{l}{}                   & $k=10^4$& 13160.80 & 3368.25 & 227.03 & 3794.34 & N/A    & 7674.99  & N/A            & 3368.25 & 12926.79 & \textbf{183.74} \\ \bottomrule
\end{tabular}
\vspace{-1em}
\end{table*}

%The impact of $k$ for \pick in terms of memory cost
We verify the following three questions: \textbf{1)} whether \pick outperforms existing algorithms for (very) large $n$ and $k$, such as $n=10^7$ and $k=10^4$; \textbf{2)} whether \pick uses less memory and performs better compared to other SOTA algorithms; and \textbf{3)} whether the proposed cost estimator in \pick shows superior accuracy in estimating runtime and memory cost.

\subsection{Experimental Settings}
\label{sec:settings}
\begin{figure*}
	\centering
\includegraphics[width=1\textwidth]{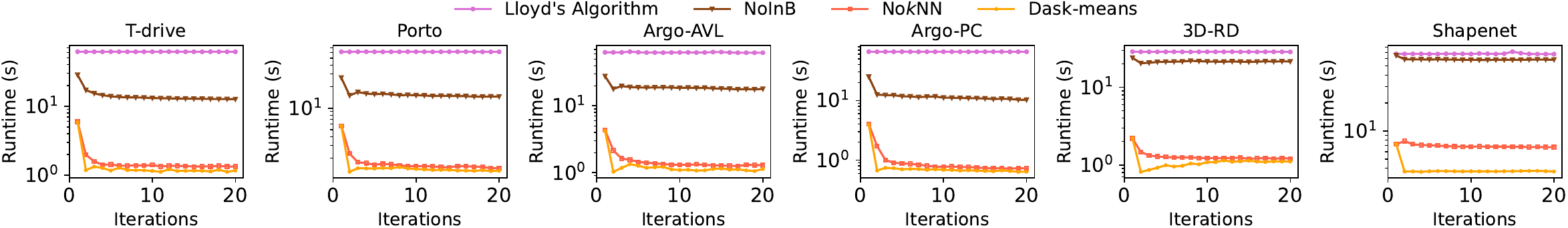}
	\vspace{-2.3em}
 \caption{The performance of $\mathtt{k}$NN and inter bound in acceleration}
\label{fig:iteration_tech}
    \vspace{-2em}
\end{figure*}

\myparagraph{Dataset}\pick is designed for spatial vectors from sensors such as GPS and lidar. For 2D datasets, we select \Tdrive \cite{yuan2010t-drive}, \Porto \cite{Proto}, and \ArgoAVL \cite{Benjamin23}, a trajectory dataset from test vehicles in a specific area. For 3D datasets, we select point cloud data including \ArgoPC \cite{Benjamin23}, \DRD \cite{3dspatialnetwork}, and \Shapenet \cite{shapenet}. We also validate our algorithm on high-dimensional datasets. The trajectory datasets are from Argoverse, denoted as \ArgoETD, and ApolloScape, referred to as \ApollTD, with each trajectory data embedded into fixed-length vectors. The details are provided in Table~\ref{tab:datasets}.

\myparagraph{Implementations}
We implement \pick and comparisons using C++. We test the performance of our algorithm on both a server and a smartphone: 1) The server, equipped with an i9-14900KF CPU and 128 GB RAM, allows us to simulate the $k$-means task on resource-limited devices and easily implement our lightweight estimator to predict runtime and memory usage; and 2) We test our algorithm on an OPPO Reno11 5G Android smartphone \cite{OPPO} equipped with a Dimensity 8200 CPU and 12 GB of RAM. Due to page limitations, the detailed information about the smartphone (see Table~\ref{tab:oppo}) and images of the \kmeans algorithms running on it (see Table~\ref{fig:oppo_running}) are presented in the Appendix~\ref{sec:appendix}. This validation demonstrates its superior performance on edge devices compared to other algorithms. Our code is publicly available on GitHub \cite{pick-repo}.

% \usepackage{multirow}
\begin{table*}[]
\centering
\ra{1}
\caption{Validating pruning power of \pick in high-dimensional datasets.}
\label{tab:high_dimension}
\vspace{-1em}
\setlength{\tabcolsep}{4.8pt}
\begin{tabular}{cccccccccccc}
\toprule
\multicolumn{1}{c}{\textbf{Dataset}}  & Settings& \texttt{Lloyd}   & \texttt{NoBound} & \texttt{Dual-tree} & \texttt{Hamerly} & \texttt{Drake}  & \texttt{Yinyang} & \texttt{Elkan}  & \texttt{NoInB} & \texttt{No$\mathtt{k}$NN} & \texttt{Dask-means} \\ \midrule
\multirow{3}{*}{\ApollTD}   
& $k=10^2$   
& 26.11  & 217.18 & 47.25  & \textbf{23.07}  & 108.68  
& 49.53  & 26.07  & 24.19  & 42.99  & 24.04 \\
& $k=10^3$  
& 258.03 & 1897.33& 158.05 & 227.64  & 1416.37  
& 491.13 & 261.01 & 40.65  & 263.91  & \textbf{40.45}  \\
& $k=10^4$ 
& 2826.21& N/A    & 1352.58& 2327.68& N/A   
& 5119.22& 2815.43& 193.18 & 2547.86& \textbf{192.69} \\ 
\midrule

\multirow{3}{*}{\ArgoETD} 
& $k=10^2$   
& 49.52  & 408.68 & 100.50 & \textbf{45.21}  & 211.17  
& 94.99  & 49.58  & 46.42  & 83.35  & 46.23  \\
& $k=10^3$ 
& 495.19 & 3700.64& 323.40 & 448.26 & 2731.09
& 954.21 & 496.54 & 78.01  & 521.65 & \textbf{77.59}  \\
& $k=10^4$   
& 5680.86&  N/A   & 2719.24& 4780.29&  N/A   
&10468.70& 5864.86& 384.28 & 5169.44& \textbf{378.82}\\
\bottomrule
\end{tabular}
\vspace{-1em}
\end{table*}
\myparagraph{Comparisons}To answer the first two questions, besides Lloyd's algorithm, we compare \texttt{Dask-means} with the most memory-efficient \kmeans algorithms including \texttt{NoBound} \cite{Xia2020} and \texttt{Hamerly} \cite{Hamerly2010}, and three widely-used algorithms, including \texttt{Dual-tree} \cite{Curtin2017}, \texttt{Drake} \cite{Drake2013}, and \texttt{Yinyang} \cite{Ding2015}. Moreover, we compare \pick with the \kmeans algorithm used in scikit-learn \cite{Pedregosa2011}, known as \texttt{Elkan} \cite{Elkan2003}.

To answer the third question, we use several SOTA cost estimators as competitors (see Section~\ref{sec:cost_pre}), including \XGBoost \cite{GunnarssonBW23}, \DisNet \cite{EggenspergerLH18}, and \AutoML \cite{MohrWTH21} to predict runtime. We configure the \XGBoost with a learning rate of 0.1 and restrict the maximum depth of each tree to 5. Additionally, it specifies that 100 trees are used in the \XGBoost model, with a column sampling ratio of 0.3 per tree. Moreover, we set up the \DisNet with two hidden layers, the first having 128 neurons and the second with 64 neurons, both of which use ReLU activation. The \DisNet model is trained for 1000 epochs with a default learning rate of 1e-4. For \AutoML, we set the regularization coefficient as 0.1 and then run the model at a maximum iteration number of 1000 times with the tolerance for convergence set as 0.1.
For memory prediction, although there are many estimation methods (see Section~\ref{sec:cost_pre}), none are designed for $k$-means tasks in resource-constrained devices.

\begin{figure*}
	\centering
\includegraphics[width=1\textwidth]{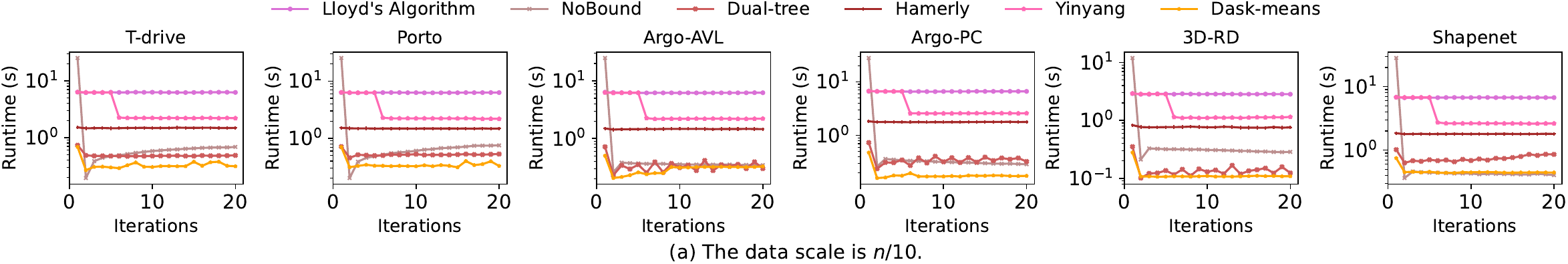}
\includegraphics[width=1\textwidth]{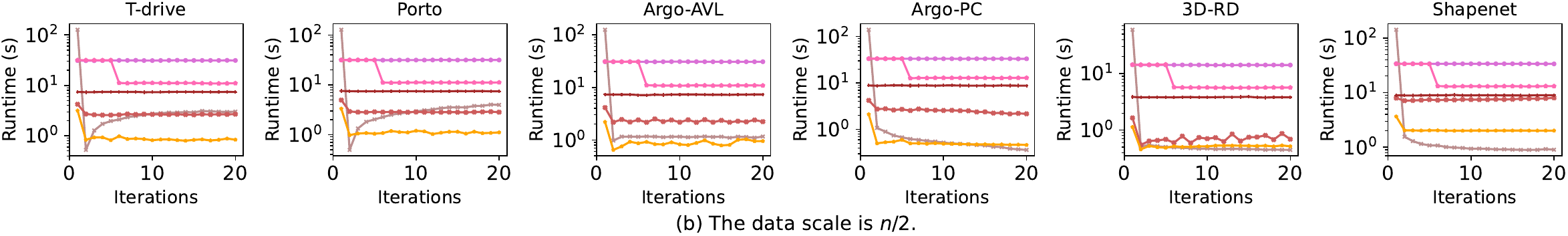}
	\vspace{-2.3em}
       \caption{The per-iteration runtime of the \kmeans algorithm under different data scales.}
	\label{fig:tech_n_2}
    \vspace{-1em}
\end{figure*}

\begin{figure*}
	\centering
\includegraphics[width=1\textwidth]{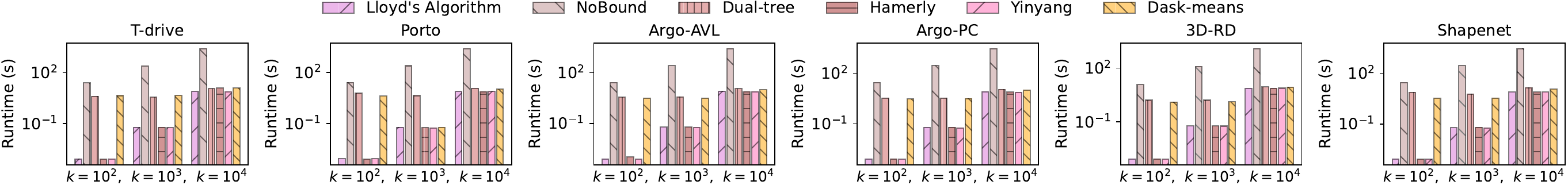}
	\vspace{-2.3em}
 
  \caption{The initialization runtime of each \kmeans algorithm.}
	\label{fig:initialization}
    \vspace{-2em}
\end{figure*}

\subsection{Efficiency of Proposed Accelerator}
\label{sec:Efficiency_of_pick}
\myparagraph{$\mathtt{k}$NN and Inter Bound's Effectiveness in Accelerating} We demonstrate the effectiveness of $\mathtt{k}$NN and the inter bound used in \pick. The algorithm only using the inter bound is called \texttt{No$\mathtt{k}$NN}, while the one only using $\mathtt{k}$NN is called \texttt{NoInB}. By default, we set the leaf node capacity to \( f = 30 \). We also limit the maximum number of iterations to 20 to save time. This is because, as depicted in Fig.~\ref{fig:iteration_tech}, each iteration's runtime has 
already stabilized after the $15$-th and $20$-th iterations.

\finding (1) Both \texttt{No$\mathtt{k}$NN} and \texttt{NoInB} can efficiently accelerate Lloyd's algorithm by pruning the number of distance computations. (2) \texttt{NoInB} exhibits higher efficiency compared to \texttt{No$\mathtt{k}$NN}. This indicates that using $\mathtt{k}$NN yields higher pruning power than using inter bound. 

%名字取的不合理Table~\ref{tab:runtime}
\myparagraph{Comparisons with SOTAs}We compare \pick with other algorithms from two aspects, including the runtime of each iteration and the total runtime of \kmeans algorithms (due to page limitations, we provide a comparison of the runtime of each iteration in Appendix~\ref{sec:Efficiency_of_pick1}). As shown in Table~\ref{tab:runtime}, we evaluate the efficiency of \pick by comparing its runtime against SOTA \kmeans algorithms.

\finding (1) When $k$ is small (e.g., $k=10^2$), \pick performs better than most SOTAs in pruning power in most cases, but it's not always the best. For example, \texttt{Elkan} outperforms it because \pick requires additional time to construct the spatial vector index and centroid index, while $\mathtt{k}$NN on these indexes is inefficient when \( k \) is small. (2) Whereas when $k$ is large, \pick demonstrates superior runtime performance due to the effective pruning power by the centroid index. For instance, when \( k = 10^4 \), \pick achieves a speedup of over 168 times compared to Lloyd's algorithm on the \ArgoPC dataset. 
(3) When \( k = 10^4 \), \texttt{Elkan} is unable to execute because it requires to store \( n \times k \) lower bounds, which leads to excessive memory cost. Similarly, \texttt{Drake} is unable to execute because it requires storing at least \(\frac{k}{8}\) lower bounds for each spatial vector, which is memory-intensive. (4) As shown in Fig.~\ref{fig:tech_n_2}, \pick demonstrates the best acceleration in almost all data scales. However, its performance diminishes with smaller data scales. This decline is attributed to the fact that, at smaller scales, our proposed \( \mathtt{k} \)NN search for spatial vectors on the index does not significantly outperform one-by-one searching, while still requiring additional time to build the index.
%NoBound is the best with respect to memory for all 12 tasks, and Dual-tree is the best with respect to time for only three tasks. This is because $\mathtt{k}$NN of \pick sometimes has weaker pruning when fewer centroids are indexed, and NoBound's only needs to store a $k \cdot k$ distance matrix.

\myparagraph{Efficiency of Initialization} As shown in Fig.~\ref{fig:initialization}, we compare the initialization times of various \( k \)-means algorithms, such as the time for building the centroid index. This comparison helps clarify that the limited acceleration effects of certain algorithms are caused by the significant time consumed during their initialization. It is worth noting that we exclude \texttt{Elkan} and \texttt{Drake} from our comparisons due to their lack of memory efficiency. At \( k = 10^4 \), their initialization processes would result in excessive memory overhead on the server.

\finding (1) The initialization time of \texttt{NoBound} is the longest because it requires computing an \( n \times d \) distance matrix, which may contribute to its inefficiency. (2) The initialization time of \pick is longer than that of Lloyd's algorithm, \texttt{Hamerly}, and \texttt{Yinyang} due to the additional time needed to build an index over spatial vectors. (3) Different values of \( k \) have a significant impact on construction time, but the initialization time of \pick is less affected.

\begin{figure}
	\centering
\includegraphics[width=0.49\textwidth]{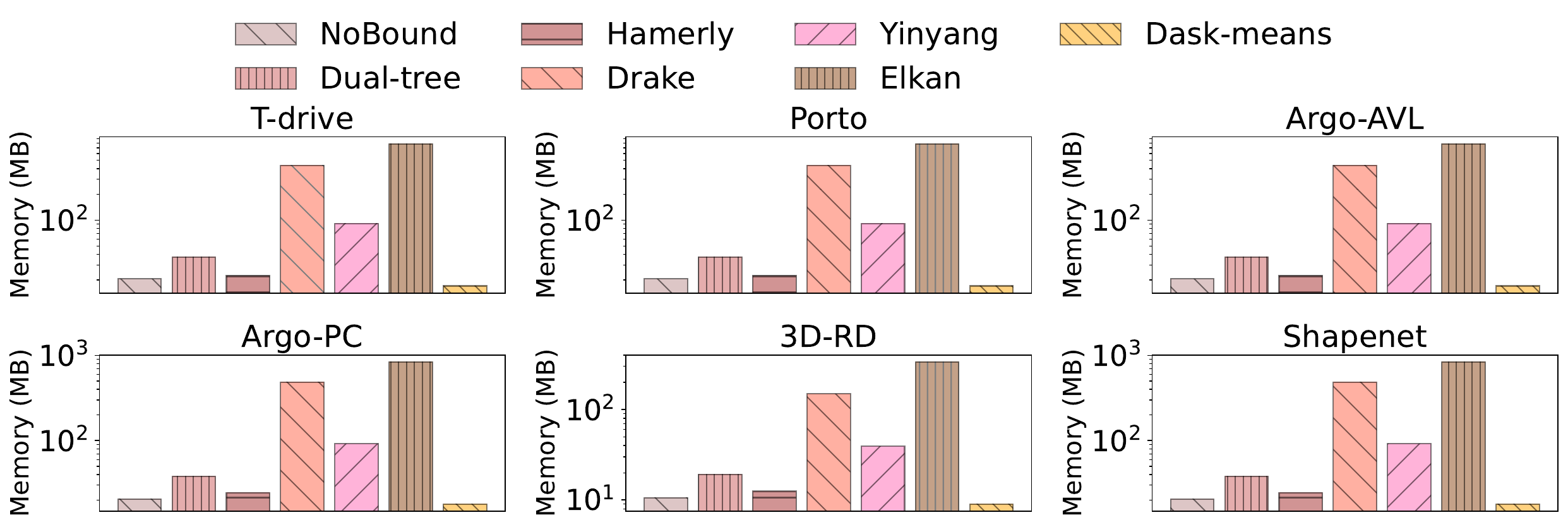}
	\vspace{-2em}
 \caption{The performance of \pick in terms of memory cost.}
	\label{fig:memory}
    \vspace{-1.5em}
\end{figure}

\myparagraph{Comparison of Space Efficiency}We compare \pick with the other SOTAs in memory cost in Fig.~\ref{fig:memory} (we set $k=10^3$). Memory cost is the amount of memory required to store the information, such as indexes and bounds in \pick.%Then, we examine the impact of varying the memory limitation on runtime.

% \usepackage{multirow}
\begin{table}[]
%\ra{1}
\centering
\setlength{\tabcolsep}{4.8pt}
\caption{Average precision of our memory estimation method.}
\label{tab:over}
\vspace{-0.7em}
\begin{tabular}{ccccc}
\toprule
\multicolumn{1}{c}{\textbf{Parameters}}            & \multicolumn{4}{c}{\textbf{Accuracy}}          \\ \midrule
\multirow{2}{*}{Increasing $k$} &  $k=10$     &  $k=10^3$  & $k=10^4$  &$k=5\times 10^4$  \\
                                & 0.963 & 0.963 & 0.963 & 0.963 \\ \midrule
\multirow{2}{*}{Increasing $n$} & $n^{'} = 0.01n$ &  $n^{'} = 0.05n$ & $n^{'} = 0.25n$ & $n^{'} = n$     \\
                                & 0.989 & 0.983 & 0.976 & 0.974 \\ \midrule
\multirow{2}{*}{Increasing $f$} & $f = 30$   & $f = 100$   & $f = 150$    & $f = 200$    \\
                                & 0.964 & 0.992 & 0.993 & 0.997 \\ \bottomrule
\end{tabular}
\vspace{-1.5em}
\end{table}

\finding (1) \texttt{Elkan} and \texttt{Drake} consume significantly more memory than other algorithms. Specifically, \texttt{Elkan} requires storing \( n \times k \) lower bounds to avoid distance computations, while \texttt{Drake} stores \(\frac{k}{8}\) to \(\frac{k}{4}\) lower bounds for each spatial vector. In contrast, \pick uses less than 1\% of the memory consumed by these algorithms. Moreover, \texttt{Yinyang} also consumes more memory than \pick because it needs to store the distance from each spatial vector to its assigned cluster. (3) Although \texttt{NoBound} uses little memory, its pruning power is much worse than \pick, as shown in Table~\ref{tab:runtime}.

\myparagraph{Verification on High-dimensional Datasets}We compare \pick with selected \kmeans algorithms on high-dimensional datasets, including \ApollTD and \ArgoETD, focusing on pruning power via runtime. The runtime performance of \pick is shown in Table~\ref{tab:high_dimension}.
% \usepackage{multirow}
\begin{table*}[]
%\ra{1.1}

\setlength{\tabcolsep}{1.2pt}
\caption{The impact of the memory limitation on \pick.}
\vspace{-0.8em}
\ra{1.3}
\label{tab:memory_limitation1}
\scalebox{0.91}{
\begin{tabular}{ccccclccclccclccclccclccc}
\toprule
\multicolumn{1}{c}{\multirow{2}{*}{\textbf{\rule{0pt}{3ex}Efficiency}}}
 & \textbf{Dataset}          & \multicolumn{3}{c}{\Tdrive} &  & \multicolumn{3}{c}{\Porto} &  & \multicolumn{3}{c}{\ArgoAVL} &  & \multicolumn{3}{c}{\ArgoPC} &  & \multicolumn{3}{c}{\DRD} &  & \multicolumn{3}{c}{\Shapenet} \\ \cmidrule{2-5} \cmidrule{7-9} \cmidrule{11-13} \cmidrule{15-17} \cmidrule{19-21} \cmidrule{23-25} 
\multicolumn{1}{l}{}

& \textbf{Available Memory (MB)} 
& 15 & 20 & 30 &  
& 15 & 20 & 30 &  
& 15 & 20 & 30 &  
& 15 & 20 & 30 &  
& 15 & 20 & 30 &  
& 15 & 20 & 30 \\ \midrule

\multirow{3}{*}{\textbf{Runtime (s)}}
& $k=10^2$
& 13.51 & 14.82 & 18.14 &  
& 15.67 & 16.59 & 19.42 &  
& 10.08 & 11.03 & 13.14 &  
& 8.04 & 9.14 & 11.14 &  
& 6.82 & 6.63 & 7.28 &  
& 32.31 & 28.01 & 28.06 \\

& $k=10^3$
& 28.86 & 25.98 & 27.15 &  
& 32.66 & 28.71 & 29.39 &  
& 25.76 & 20.80 & 21.87 &  
& 17.06 & 15.74 & 16.46 &  
& 20.87 & 21.74 & 18.38  &  
& 76.95 & 93.12 & 122.15 \\

& $k=10^4$
& 105.32& 92.23  & 83.77  &  
& 117.40& 100.55 & 89.50  &  
& 86.72 & 80.61  & 67.12  &  
& 71.70 & 59.03  & 49.87  &  
& 48.41 & 54.66  & 68.39  &  
& 179.13 & 188.42  & 199.32  \\ \midrule

\multirow{3}{*}{\textbf{Pruned Vectors (M)}}
& $k=10^2$
& 18.62 & 19.32 & 19.66 &  
& 18.37 & 19.15 & 19.57 &  
& 18.12 & 19.06 & 19.55 &  
& 19.35 & 19.68 & 19.84 &  
& 6.82 & 6.90 & 7.62 &  
& 8.92 & 12.20 & 15.04       \\

& $k=10^3$
& 15.91  & 17.85 & 18.92 &  
& 15.07  & 17.36 & 18.63 &  
& 13.72  & 16.80 & 18.49 &  
& 16.20  & 18.11 & 19.19 &  
& 1.98   & 3.91 & 5.79  &  
& 2.30   & 5.37   & 9.37  \\

& $k=10^4$
& 7.24   & 12.18 & 15.47 &  
& 5.50   & 10.61 & 14.51 &  
& 5.16   & 10.73 & 15.12 &  
& 5.10   & 11.41 & 15.86 &  
& 0.09   & 0.88 & 2.71  &  
& 0.02   & 0.71   & 3.74   \\ \bottomrule
\end{tabular}}
\vspace{-1em}
\end{table*}

\finding \pick performs the best in most cases. However, when tackling high-dimensional datasets, almost all \kmeans algorithms perform poorly. This is due to the ``curse of dimensionality''. It is worth noting that the acceleration performance of \pick is significantly lower in high-dimensional cases compared to low-dimensional ones. For example, it is only about 15 times faster than Lloyd's algorithm.

\myparagraph{Verification on Edge Devices} We validate \pick on a smartphone and compare its runtime with SOTAs. Due to the maximum response time limits imposed by the Android environment on program execution, the data scale is set to $\frac{1}{20}$ of the original dataset, with $k=100$.

\begin{figure}
	\centering
\includegraphics[width=0.49\textwidth]{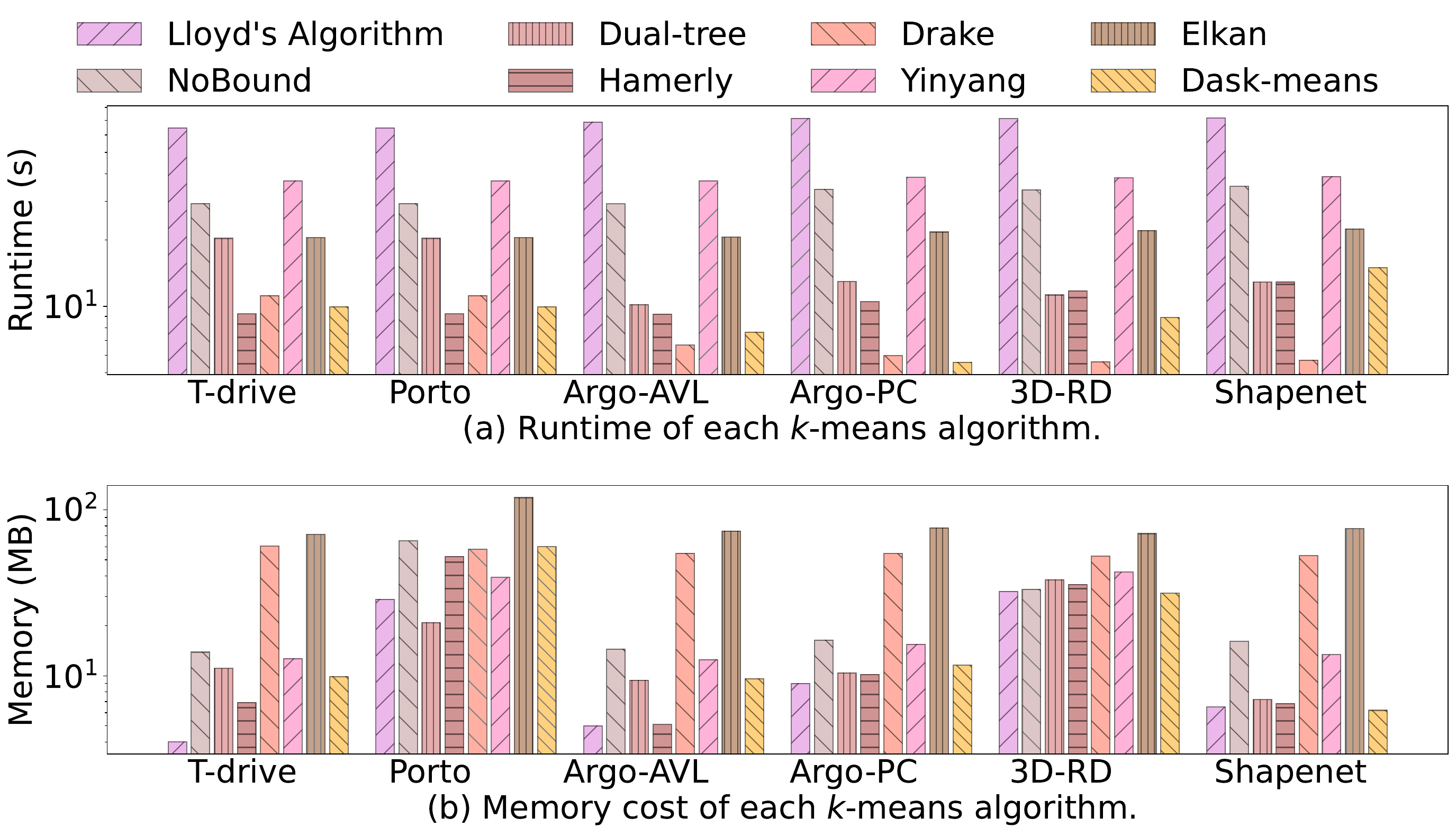}
	\vspace{-2em}
 \caption{The performance of \pick in the smartphone.}
	\label{fig:edge}
    \vspace{-2em}
\end{figure}

\finding (1) As shown in Fig.~\ref{fig:edge}(a), \pick is generally very fast, although it can be slower than \texttt{Drake} in some cases. However, \texttt{Drake} needs to store between $\frac{k}{8}$ and $\frac{k}{4}$ lower bounds for each spatial vector, which consumes significantly more memory than \pick, making it not memory-efficient. (2) As shown in Fig.~\ref{fig:edge}(b), in some cases, \pick consumes more memory than \texttt{Hamerly}, as \texttt{Hamerly} only requires storing one upper bound and one lower bound for each spatial vector. However, its pruning power is weaker compared to \pick.

\myparagraph{Summary of Lessons Learned} Through the evaluation of \pick in runtime and memory cost, we further learn:

%不要用定冠词
\begin{itemize}%\setlength\itemsep{-0.2em}
    \item Both \texttt{No}$\mathtt{k}$\texttt{NN} and \texttt{NoInB} accelerate Lloyd's algorithm, but \texttt{NoInB} is much more efficient, likely because our estimated bounds are too loose.
    
    \item The value of $k$ has only a slight effect on efficiency. This is consistent with the observation that $\log_2k$ and the dataset scale $n$ have a linear relation with the running time.

    \item For high-dimensional datasets, \pick can still accelerate Lloyd's Algorithm; however, its acceleration performance is significantly lower than that for low-dimensional datasets due to the ``curse of dimensionality''.
    %Reducing available memory increases $f$, ensuring lower memory cost. However, as $f$ increases, performance degrades since pruning with a larger radius $r$ has a lower success probability.
\end{itemize}

\subsection{Evaluation of Our Cost Estimator}
\label{sec:exp_cost}
We test our cost estimator to demonstrate its superiority in predicting memory cost and runtime. We generate 2000 $k$-means tasks as a sample set and divide them into three parts: 80\% for training, 10\% for validation, and 10\% for testing. For each $k$-means task, we randomly select a dataset with a size ranging from $1 \times 10^5$ to $1 \times 10^8$ and choose $k$ randomly between $1 \times 10^2$ and $1 \times 10^4$. We then extract the features, run \pick, and record the runtime. It is important to note that for predicting runtime, we choose \(\beta = 4\) and \(\sigma = 50\) as default parameters (details on the selection of a suitable \(\beta\) and \(\sigma\) can be found in Appendix \ref{sec:Selection}).

\begin{figure*}
	\centering
\includegraphics[width=1\textwidth]{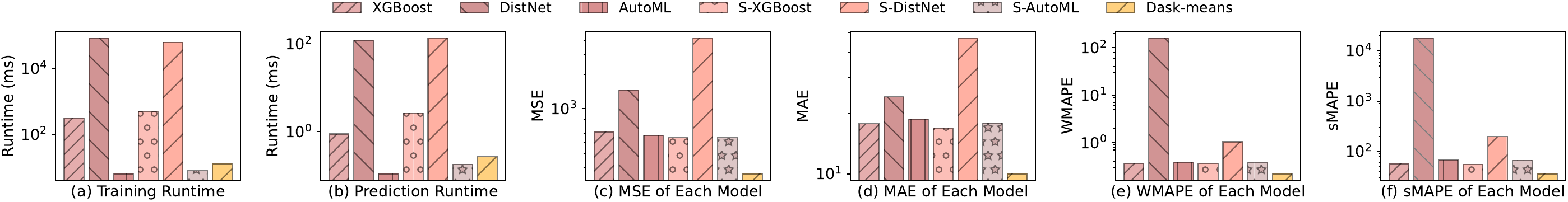}
	\vspace{-1.8em}
 \caption{The performance of our cost estimator in terms of predicting runtime.}
	\label{fig:cost_accuracy}
    \vspace{-1.7em}
\end{figure*}
\myparagraph{Memory Cost Estimation}
We first show that the proposed cost estimator can accurately estimate the memory cost of \pick. It is worth noting that the estimated memory cost is often less than the actual memory used (see Section~\ref{sec:memory}). Hence, we measure the accuracy of our memory estimation method using the ratio of the estimated memory to the actual memory consumed. 

\finding As shown in Table~\ref{tab:over}, when $k$ (i.e., the number of centroids) increases, the prediction accuracy of our proposed cost estimator remains unchanged. This is because the memory used by the centroid index is much smaller than the memory used to construct the spatial vector index. Moreover, as $n^{'}$ (the number of spatial vectors for \kmeans) increases, the prediction accuracy of our proposed cost estimator decreases. This is because an increase in dataset scale leads to more nodes in index structure, and we estimate memory usage by assuming that each index node only includes the spatial vector and pointer it stores, without considering additional information like locks in the ``vector'' structure. Hence, increasing the number of nodes adds more unestimated information, reducing the accuracy of the memory estimation. Similarly, when $f$ increases, the index has fewer nodes, resulting in higher prediction accuracy.

\myparagraph{Impact of Memory Constraints} 
As shown in Table \ref{tab:memory_limitation1}, under various memory limits, we evaluate the efficiency of the memory-tunable index for accelerating $k$-means tasks.

\finding (1) As shown in Table \ref{tab:memory_limitation1}, as memory cost increases, the number of pruned spatial vectors also rises. A higher memory cost leads to a smaller $f$, and $\mathtt{k}$NN search on an index with a smaller $f$ consistently results in a reduced search radius. Consequently, more unnecessary spatial vectors and nodes are pruned, improving the index's pruning capability. (2) We find that as memory increases, the runtime does not necessarily decrease. This is because, while more memory improves the index's pruning power, it also requires additional time to build the index, which offsets the time saved from improved pruning. Additionally, as $k$ increases, the runtime also increases, indicating that $k$-means converge faster with smaller values of $k$.

%\myparagraph{Evaluation of cost estimator}
%We evaluated the accuracy of predicting memory and runtime for \kmeans. Specifically, we first test whether our model achieves the best accuracy. Next, we verify that our model can refine the predicted runtime. Finally, we confirm that our model can accurately predict the memory usage of \kmeans.

\myparagraph{Comparisons with SOTAs in Runtime Prediction} We use four metrics \cite{ChiccoWJ21} to assess the accuracy in terms of runtime: Mean Squared Error (\textbf{MSE}), Mean Absolute Error (\textbf{MAE}), Weighted Absolute Mean Percentage Error (\textbf{WAMPE}), and Symmetric Mean Absolute Percentage Error (\textbf{SMAPE}). Then we compare our cost estimator with SOTA models, observing each model's training time, prediction time, and accuracy in predicting runtime. Moreover, we modify existing models to predict each iteration separately and then sum the predictions to obtain the total runtime. The modified models are labeled with \texttt{S-}, such as \texttt{S-}\XGBoost, \texttt{S-}\DisNet, and \texttt{S-}\AutoML.

\finding (1) Fig.~\ref{fig:cost_accuracy}(a) shows that our cost estimator has the shortest training time compared to others, similar to \AutoML. This is because both the proposed cost estimator and \AutoML require only one pass through the dataset to obtain regression parameters. (2) Fig.~\ref{fig:cost_accuracy}(b) illustrates that prediction methods like \pick and \AutoML have similar prediction times, typically a few milliseconds. Additionally, compared to the overall runtime of \pick, which requires several seconds to minutes per iteration, this prediction time is negligible. (3) Fig.~\ref{fig:cost_accuracy}(c), (d), (e), and (f) demonstrate that our cost estimator achieves the highest prediction accuracy, with the smallest MSE, MAE, WMAPE, and sMAPE compared to others. Moreover, it shows that using complex iterative algorithms does not necessarily lead to better performance. For example, regression models often achieve higher accuracy than \XGBoost. Moreover, models such as \XGBoost perform worse after modification.

\myparagraph{Summary of Lessons Learned} Through the evaluation of our cost estimator, we further learn:
\begin{itemize}%\setlength\itemsep{-0.2em}
    \item As the leaf node capacity $f$ increases, the runtime of the $k$-means task does not necessarily increase. This is because, although pruning with a larger radius $r$ has a lower success probability, the time to build the index also decreases.
    
    \item Our cost estimator predicts runtime more accurately than others. However, it's important to note that the runtime of different $k$-means tasks varies significantly, leading to discrepancies between predicted and actual times that can be several times the actual $k$-means runtime.

    \item Our runtime adjustment method dynamically corrects runtime. However, if parameters like $\sigma$ are not chosen properly, such as $\sigma = 2$, its adjustment capability will significantly decrease and may decrease prediction accuracy.
\end{itemize}

\section{Conclusions}
To accelerate \kmeans for simplifying large-scale spatial vectors, we leveraged fast $\mathtt{k}$NN search and assigned spatial vectors to the nearest centroid in batches by indexing on both spatial vectors and centroids. Without updating the bounds for the next iteration, novel bounds were designed to further accelerate the $\mathtt{k}$NN search. Moreover, we designed a lightweight cost estimator to predict the $k$-means memory cost and runtime accurately. Experiments on real-world datasets verified the efficiency of \pick on resource-constrained devices. 

In future work, we will design a distributed \kmeans on resource-constrained devices to leverage the remaining computational power of edge devices to accelerate \kmeans. Additionally, we plan to design a more lightweight and accurate cost estimator and extend it to other iterative algorithms.

% or



% % use section* for acknowledgment
% \ifCLASSOPTIONcompsoc
%   % The Computer Society usually uses the plural form
%   \section*{Acknowledgments}
% \else
%   % regular IEEE prefers the singular form
% \fi



%\balance
\ifCLASSOPTIONcaptionsoff
  \newpage
\fi
\newpage
\bibliographystyle{abbrv}%
\bibliography{references}

\clearpage

\section{Appendix}
\label{sec:appendix}
\subsection{Complexity Analysis}
\label{sec:Complexity Analysis}
We analyze the time complexity of the proposed pruning mechanism.
We first analyze the construction time and search time on different types of indexes (using Ball-tree structures). A balanced Ball-tree containing $n$ spatial vectors has a height of $\lceil \log_2\frac{2n}{f} \rceil$ when each leaf node contains $\frac{f}{2}$ spatial vectors. 
Assume that the dataset consists of $d$-dimensional spatial vector.
Then the construction time of a balanced Ball-tree is $O(dn\log_2\frac{2n}{f})$ \cite{Omohundro1989} and the $\mathtt{k}$NN search on a balanced Ball-tree costs $O(d(\log_2\frac{2n}{f}+f))$ time, which is the best case. On the other hand, for a degenerate Ball-tree with height $n-f$, the construction time is $O(dn^2)$ and the complexity of $\mathtt{k}$NN search can be as high as $O(dn)$ in the worst case.

In each iteration of the clustering algorithm, it takes $O(dk\log_2\frac{2k}{f})\sim O(dk^2)$ time to create a Ball-tree on $C$. Then, in lines 5-7, the computation of the inter bound for each centroid costs $O(dk(\log_2\frac{2k}{f}+f))\sim O(dk^2)$ time. The {\textbf{Assign}} function, in the worst case, needs to scan the whole Ball-tree on $\mathbf{D}$ and this process costs $O(dn(\log_2\frac{2k}{f}+f))\sim O(dnk)$ time. 
%\songsong{maybe we need to consider the best case.} 
Lastly, it takes $O(k)$ time to refine centroids. Thus, the total time complexity of \pick is $O(d(n+2k)\log_2\frac{2k}{f}+d(n+k)f)\sim O(d(n+2k)k)$. 

Note that the total runtime is related to the iteration number of $k$-means. However, the above time complexity for each iteration is just theoretical analysis, and calculating the total runtime is still challenging, as it is not clear when $k$-means tasks converge. Next, we will design a cost estimator to predict the memory cost and the runtime accurately for $k$-means tasks.

\subsection{Additional Comparisons with SOTAs}%重复More Experiments on 
\begin{figure}
	\centering
\includegraphics[width=0.49\textwidth]{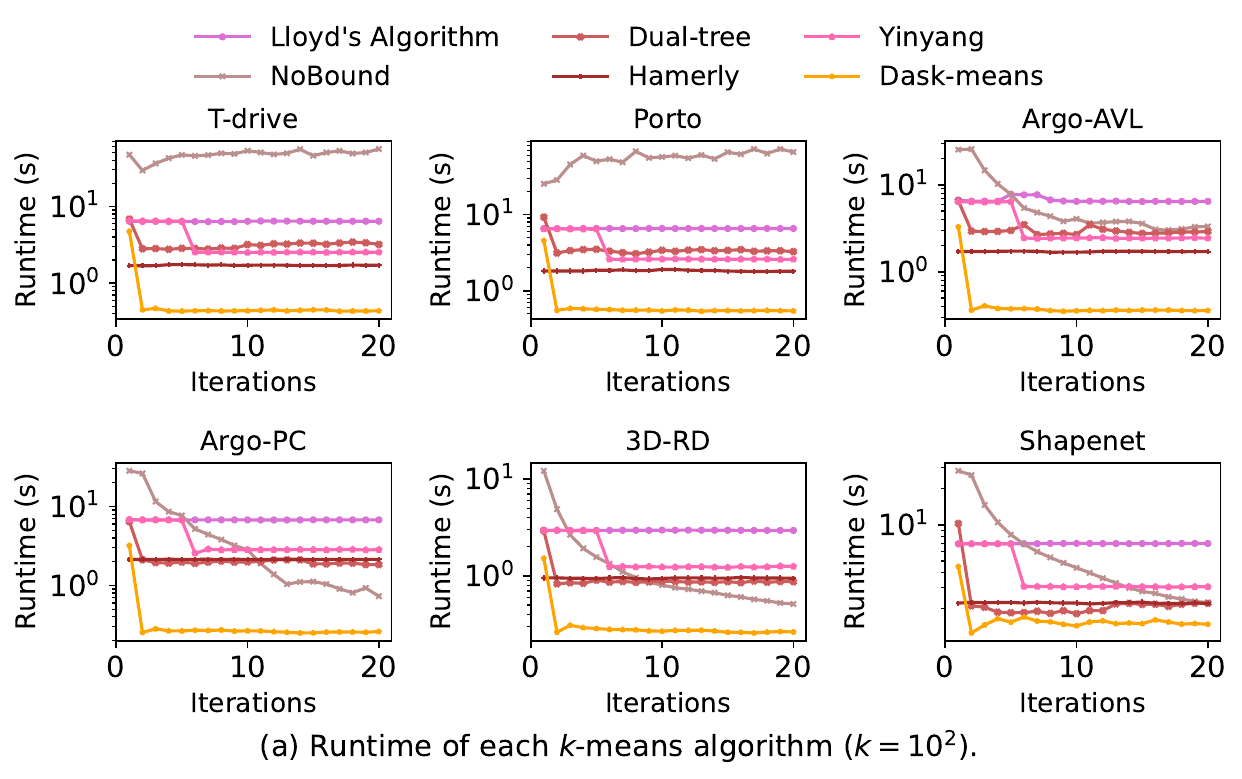}
\includegraphics[width=0.49\textwidth]{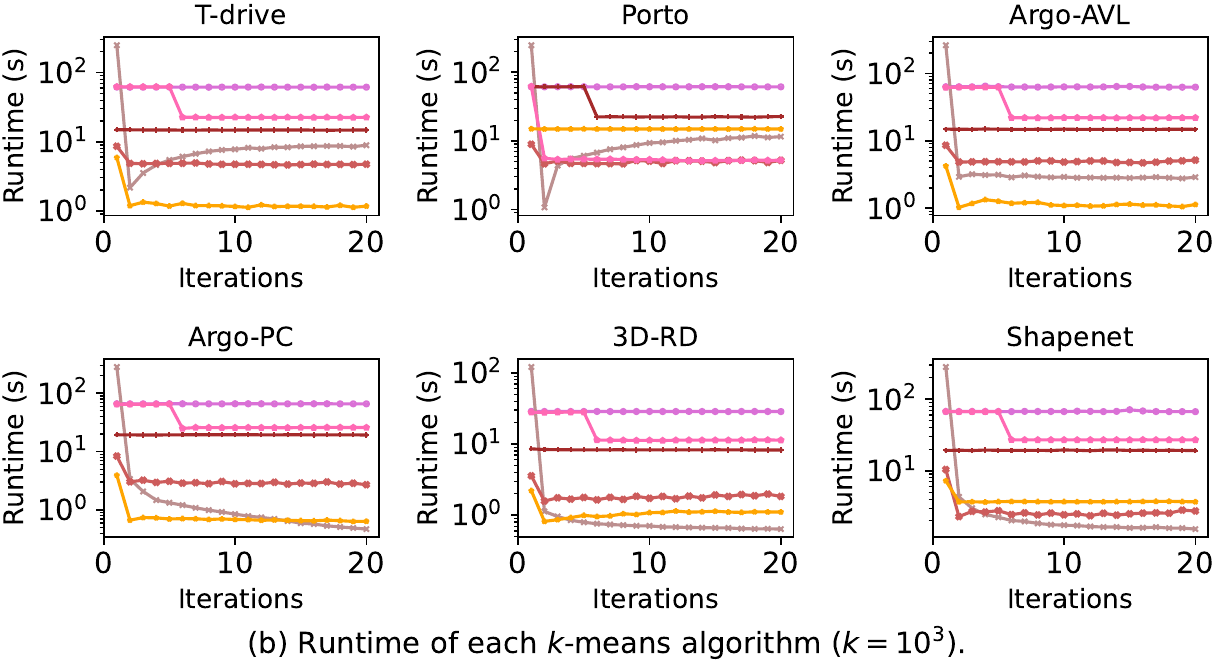}
\includegraphics[width=0.49\textwidth]{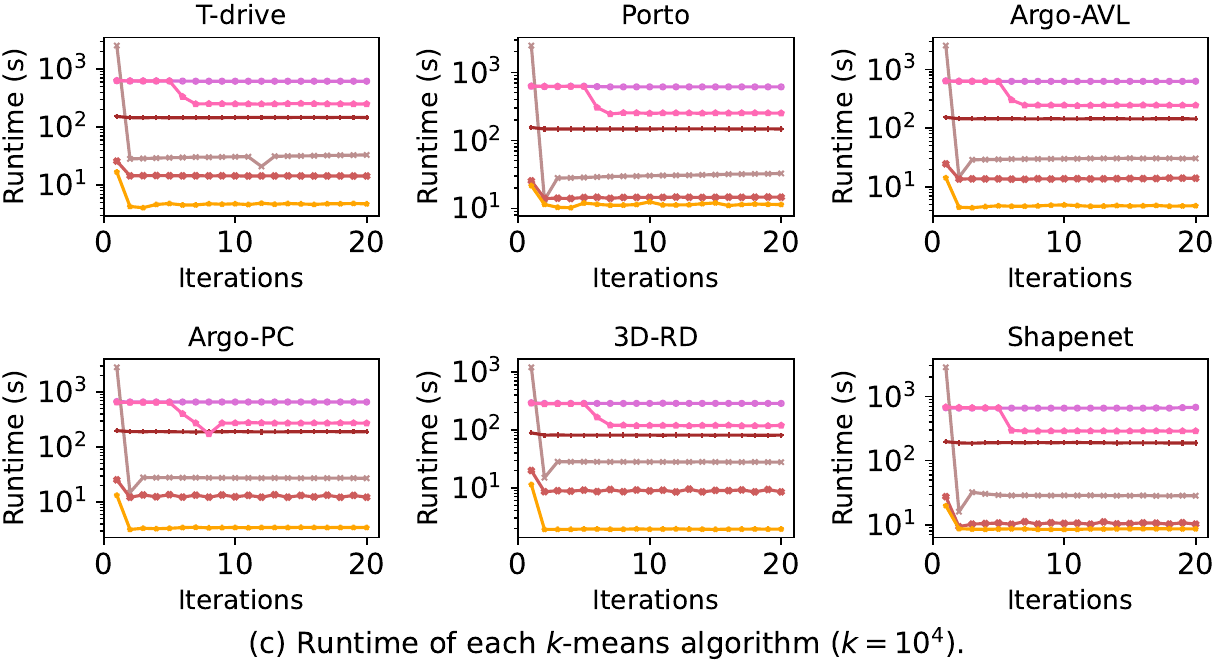}
	\vspace{-2.3em}
 
       \caption{The runtime performance of \kmeans algorithms in each iteration.}
	\label{fig:iteration_tech100}
    \vspace{-1.5em}
\end{figure}

\label{sec:Efficiency_of_pick1}
As shown in Fig.~\ref{fig:iteration_tech100}, we evaluate the efficiency of \pick by comparing its per-iteration runtime with other SOTA \kmeans algorithms.

\finding (1) \pick achieves the best per-iteration acceleration in most cases when \( k \) takes on different values. However, when \( k \) is not large, such as $k=10^3$, as shown in \Porto in Fig.~\ref{fig:iteration_tech100}(b), \pick is slower than \texttt{NoBound}, \texttt{Hamerly}, and \texttt{Dual-tree}. This is because \pick incurs additional time due to constructing two extra indexes, while its pruning power is less effective. Moreover, we observe that \texttt{NoBound} does not accelerate Lloyd's algorithm and is even slower when \( k \) is small, as shown in \Tdrive in Fig.~\ref{fig:iteration_tech100}(b). (2) When \( k \) is relatively large, such as \( k = 10^4 \), the per-iteration runtime stabilizes after the first five rounds. (3) The per-iteration runtime of \texttt{Hamerly} remains consistent, indicating that the pruning power from assigning each point upper and lower bounds remains stable.

\subsection{Parameter Selection for Our Cost Model}
\label{sec:Selection}
We test different $\beta$ (see Section \ref{sec:regressor}) values within the range $(1, 6)$ and various $\sigma$ values (see Section \ref{sec:GP}) within the range $(1, 100)$ to determine the suitable $\beta$ and $\sigma$. Moreover, we verify whether the interaction features can improve the prediction accuracy of the runtime prediction method.
% \usepackage{multirow}
\begin{table}[]
%\ra{1.2}
%\vspace{-1em}
\setlength{\tabcolsep}{1.7pt}
\caption{The impact of the Interaction features and $\beta$.}
\label{tab:beta}
\vspace{-0.8em}
\begin{tabular}{ccccclcccc}
\toprule
\multirow{2}{*}{\raisebox{-0.5\height}{\textbf{Degree}}} & \multicolumn{4}{c}{\textbf{Basic Feature}} &  & \multicolumn{4}{c}{\textbf{Interaction Feature}}                       \\ \cmidrule{2-5} \cmidrule{7-10} 
                                 & MSE       & MAE      & WMAPE    & sMAPE    &  & MSE             & MAE           & WMAPE         & sMAPE          \\ \midrule
$\beta$ = 1                       & 600.48    & 18.44    & 0.41     & 62.15    &  & 525.33          & 17.39         & 0.39          & 59.06          \\
$\beta$ = 2                       & 245.62    & 12.12    & 0.27     & 37.79    &  & 229.01          & 10.52         & 0.23          & 37.44          \\
$\beta$ = 3                       & 324.07    & 11.26    & 0.25     & 31.52    &  & 264.76          & 10.01         & 0.22          & 35.72          \\
$\beta$ = 4                       & 324.68    & 11.29    & 0.25     & 28.78    &  & \textbf{227.47} & \textbf{9.44} & \textbf{0.21} & \textbf{25.72} \\
$\beta$ = 5                       & 335.36    & 12.07    & 0.27     & 34.04    &  & 232.52          & 10.75         & 0.24          & 36.90          \\
$\beta$ = 6                       & 383.38    & 13.70    & 0.30     & 40.51    &  & 1167.00         & 13.66         & 0.30          & 39.20          \\ \bottomrule
\end{tabular}
\vspace{-2em}
\end{table}

\begin{figure}
	\centering
 \vspace{-1.8em}
\includegraphics[width=0.48\textwidth]{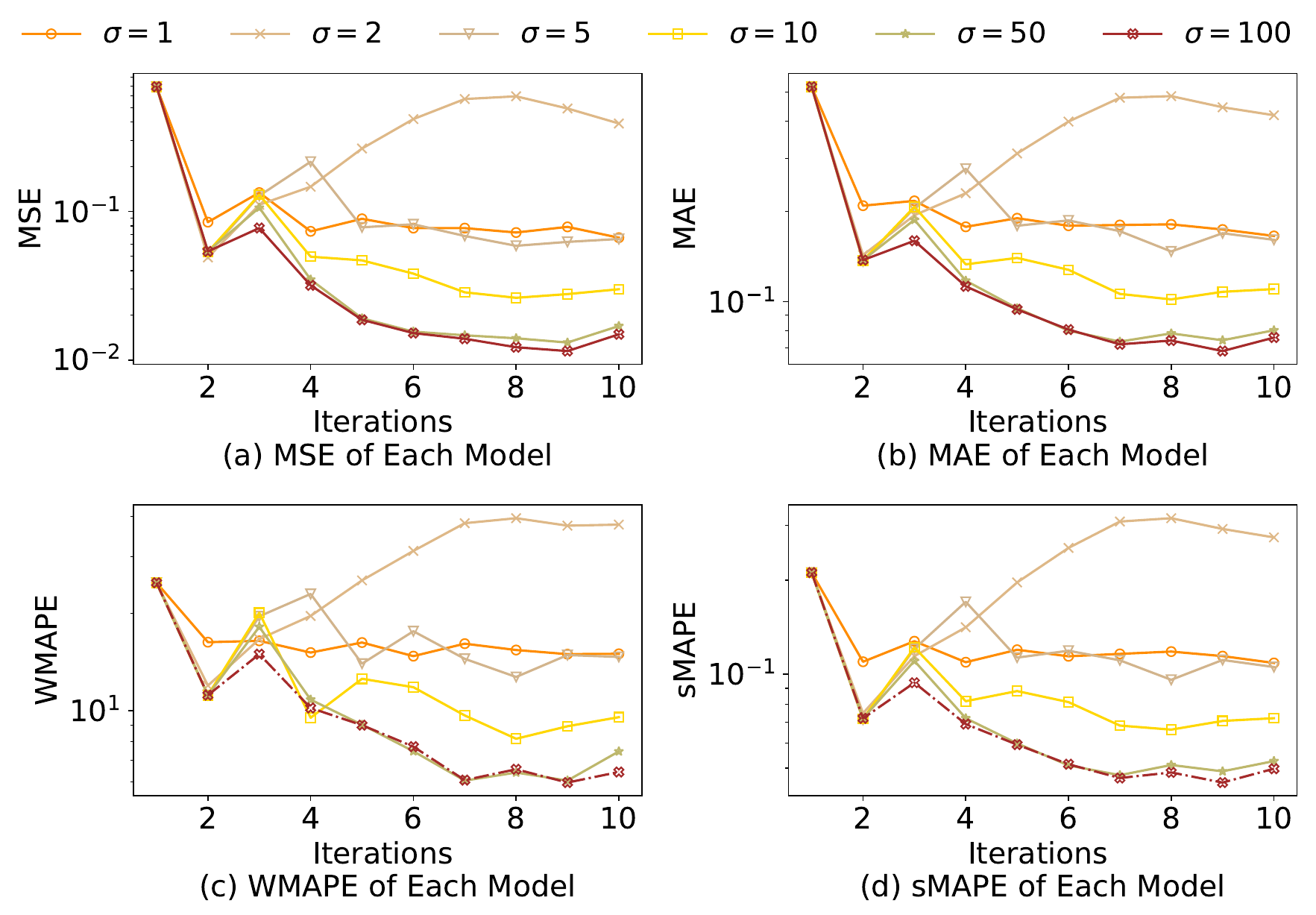}
\vspace{-1em}
 \caption{The impact of $\alpha$ in adjusting the runtime.}
	\label{fig:sigma}
    \vspace{-1em}
\end{figure}

\finding (1) As shown in Table \ref{tab:beta}, the four evaluation metrics decrease as $\beta$ increases, reaching their minimum at $\beta = 4$. Beyond this point, the metrics increase as $\beta$ continues to grow. Hence $\beta = 4$ is a suitable choice. Moreover, adding the interaction features improves the cost estimator's prediction accuracy. (2) As shown in Fig.~\ref{fig:sigma}, when \(\sigma=50\), our method reaches its strongest adjustment capability. However, if \(\sigma\) is poorly chosen, the values of the four metrics become large. For example, \(\sigma=2\) assumes a weak correlation between iterations, which is unrealistic. For example, once the final centroids are found and $k$-means is completed, there are no further iterations (the runtime for the next iteration is 0). Moreover, as runtime progresses, we find that the MSE, MAE, WMAPE, and sMAPE decrease at a roughly constant rate, indicating that adjusting $\sigma$ has less impact as the $k$-means tasks approach convergence.

\subsection{Verification for Predicted Runtime Adjustment} We verify that using the proposed cost estimator can adjust the runtime dynamically based on the posterior information they acquired from the current iteration. The calculation of metrics is obtained by comparing the predicted runtime with the actual runtime for each specified iteration. Notably, \pick without applying GP is referred to as \texttt{NoGP}.

\begin{figure}
	\centering
\includegraphics[width=0.49\textwidth]{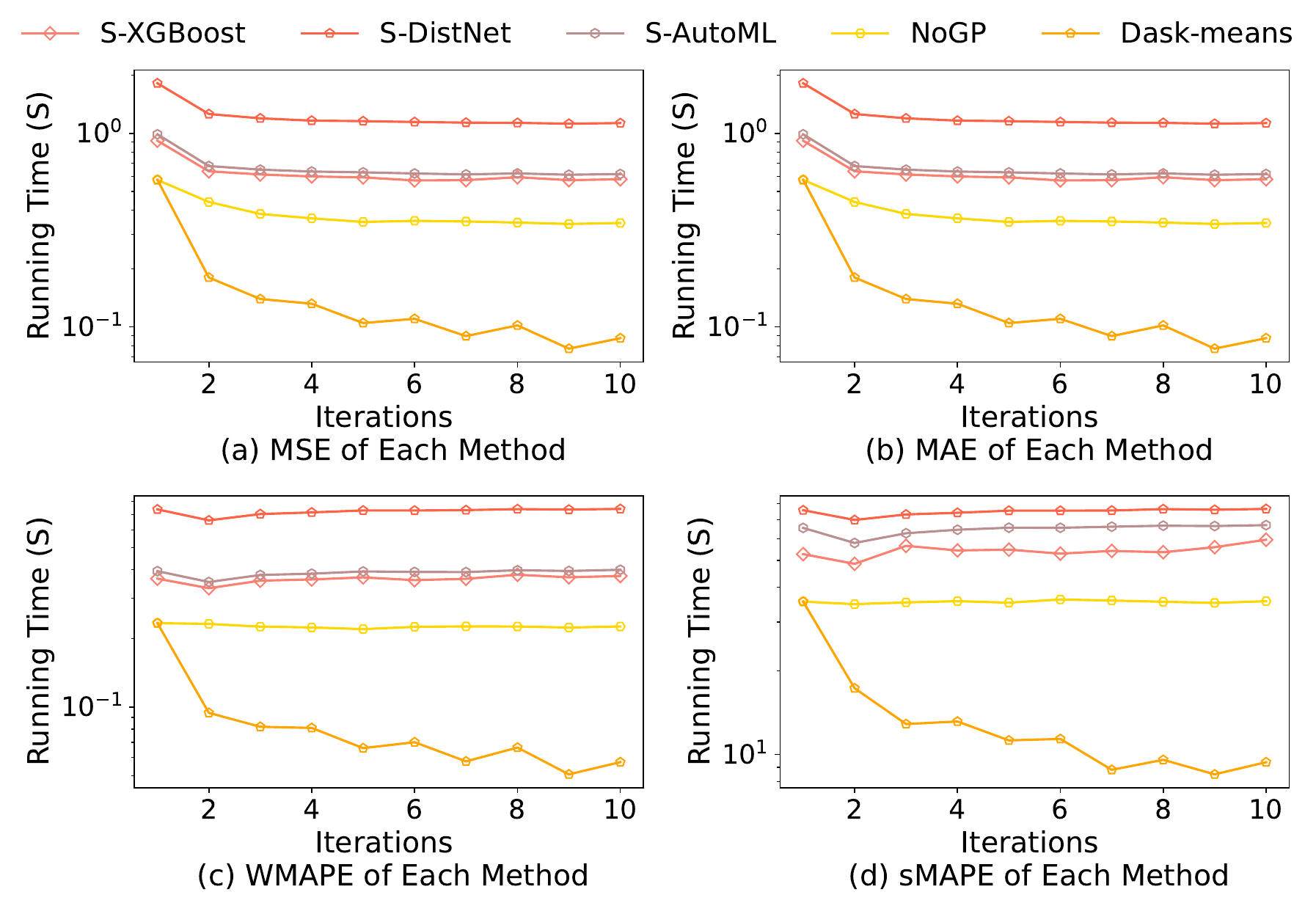}
	\vspace{-2em}
 \caption{The performance of our cost estimator in adjusting predicted runtime.}
	\label{fig:MSE}
    \vspace{-2em}
\end{figure}

\finding As shown in Fig.~\ref{fig:MSE}, compared to other SOTA methods, our cost estimator performs best across four metrics. Moreover, our cost estimator effectively corrects predicted runtime compared to \texttt{NoGP}. Furthermore, as \kmeans runs longer (with more iterations), more posterior information is obtained, improving the ability to adjust prediction times.

\begin{table}[]
\vspace{-10em}
\ra{1.1}
\centering
\setlength{\tabcolsep}{6pt}
%\setlength{\abovecaptionskip}{1em} 
%\vspace{-1em}
\caption{Information of smartphone.}
\label{tab:oppo}
\vspace{-1em}
\scalebox{1.03}{\begin{tabular}{cc}  
\toprule   
\textbf{Attribute} & \textbf{Specification}  \\  
\midrule
Model   & OPPO Reno11 5G\\
Dimensions  &  $74.3 \times 162.4 \times 7.99$ mm \\
Weight      &  182g \\
SoC &  MediaTek Dimensity 7050 (MT6877V)\\
CPU & 8-core ARM Cortex-A78/A55 (2.6/2.0 GHz) \\
GPU & ARM Mali-G68 MC4, 950 MHz, Cores: 4 \\
RAM & 12 GB, 2133 MHz \\
Storage & 256 GB \\
Display & 6.7 in, OLED, 1080 x 2412 pixels, 30 bit\\
Battery & 55000mAh/19.45Wh\\
Fast Charge & SUPERVOOCTM 67W and SUPERVOOCTM 2.0\\
Biometrics & Fingerprint and Facial Recognition\\
OS  & ColorOS 14 (Android 14)\\
Camera  & 9280 $\times$ 6920 pixels, 3840 $\times$ 2160 pixels, 30 fps\\
SIM card& Nano-SlM\\
USB & 2.0, USB Type-C\\
Bluetooth   & 5.3\\
Positioning & GPS, A-GPS, GLONASS, BeiDou, Galileo, QZss\\
\bottomrule  
\end{tabular}}
\vspace{-1em}
\end{table}
\begin{figure}
	\centering
       \vspace{-13em}
      \includegraphics[width=0.47\textwidth]{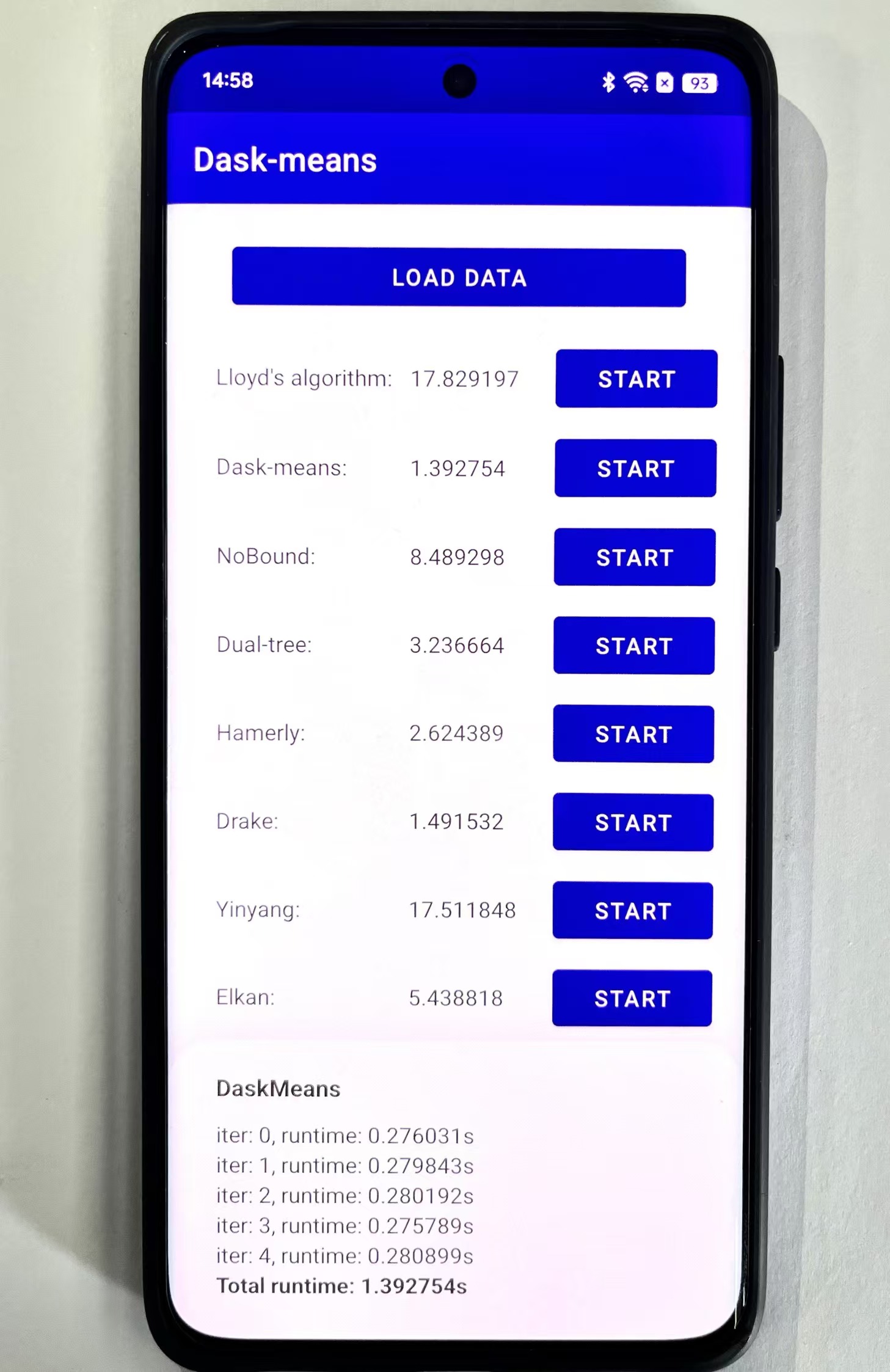}
	\vspace{-1em}
 \caption{Running \kmeans algorithms on the smartphone.}
	\label{fig:oppo_running}
    \vspace{-1em}
\end{figure}

\end{document}